\documentclass[10pt,twocolumn,letterpaper]{article}

\usepackage{titling}

\usepackage[pagenumbers]{cvpr} 

\usepackage{graphicx}
\graphicspath{{assets/}} 
\usepackage{amsmath}
\usepackage{amssymb}
\usepackage{booktabs}
\usepackage{enumitem}
\usepackage{algorithm}
\usepackage{algpseudocode}
\usepackage{bm}
\usepackage{multirow}
\usepackage{caption}
\usepackage{subcaption}
\usepackage{afterpage}

\usepackage[pagebackref,breaklinks,colorlinks]{hyperref}

\usepackage[capitalize]{cleveref}
\crefname{section}{Sec.}{Secs.}
\Crefname{section}{Section}{Sections}
\Crefname{table}{Table}{Tables}
\crefname{table}{Tab.}{Tabs.}

\def\cvprPaperID{} 
\def\confName{CVPR}
\def\confYear{2023}

\begin{document}

\title{High-Fidelity Guided Image Synthesis with Latent Diffusion Models}

\author{
Jaskirat Singh$^{1}$ \qquad Stephen Gould$^{1,2}$  \qquad  Liang Zheng$^{1,2}$\\
$^1$The Australian National University 
\qquad $^2$Australian Centre for Robotic Vision
\\
{\tt\small \{jaskirat.singh, stephen.gould, liang.zheng\}@anu.edu.au}
}

\twocolumn[{
\maketitle
\begin{center}
    \vskip -0.2in
    \captionsetup{type=figure}
    \includegraphics[width=0.977\linewidth]{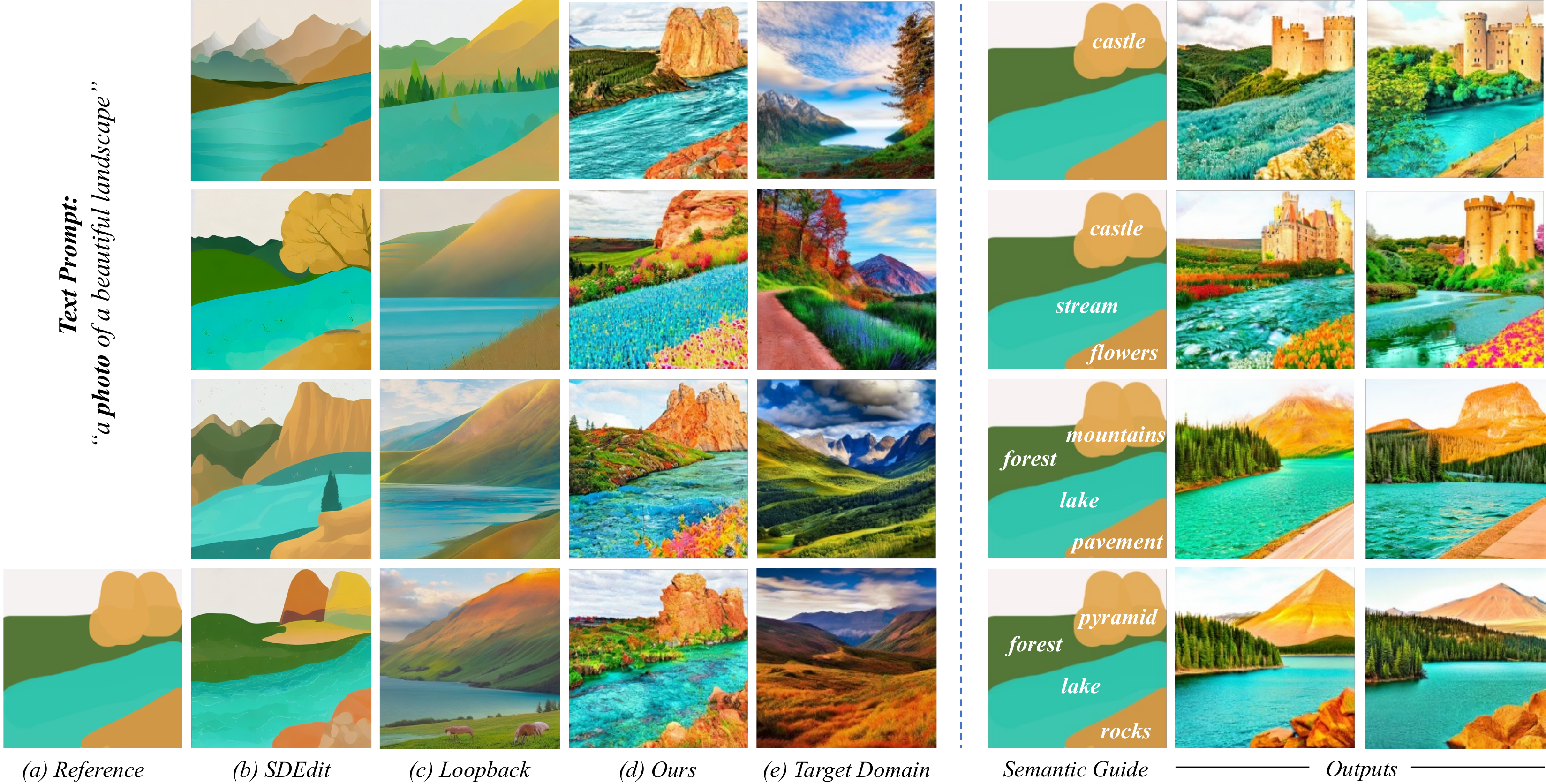}
    \vskip -0.05in
    \captionof{figure}{\emph{\textbf{Overview}}. We propose a novel stroke based guided image synthesis framework which \emph{(Left)} resolves the intrinsic domain shift problem in prior works (b), wherein 
    the final images lack details and often resemble simplistic 
    representations of the
     target domain (e) (generated using only text-conditioning).
    Iteratively reperforming the guided synthesis with the generated outputs (c) seems to improve realism but it is expensive and the generated outputs might lose faithfulness with the reference (a) with each iteration. \emph{(Right)} Additionally, the user is also able to specify the semantics of different painted regions without requiring any additional training or finetuning.}
     \label{fig:overview}
\end{center}
}]

\begin{abstract}
   \vskip -0.1in
   Controllable image synthesis with user scribbles 
   has gained huge public interest with the recent advent of text-conditioned latent diffusion models.
   The user scribbles control the color composition while the text prompt provides control over the overall image semantics. 
   However,  
   we note that prior works in this direction suffer from an intrinsic domain shift problem wherein the generated outputs often lack details and resemble simplistic representations of the target domain. In this paper, we propose a novel guided image synthesis framework, which addresses this problem by modelling the output image as the solution of a constrained optimization problem. We show that while computing an exact solution to the optimization is infeasible, an approximation of the same can be achieved while just requiring a single pass of the reverse diffusion process. 
   Additionally, we show that by simply defining a cross-attention based correspondence between the input text tokens and the user stroke-painting, the user is also able to control the semantics of different painted regions without requiring 
   any
   conditional training or finetuning. 
   Human user study results show that the proposed approach outperforms the previous state-of-the-art by over 85.32\% on the overall user satisfaction scores. 
   Project page for our paper is available at
   \url{https://1jsingh.github.io/gradop}.
\end{abstract}

\vskip -0.1in
\section{Introduction}
\label{sec:intro}

Guided image synthesis with user scribbles has gained widespread public attention with the recent advent of large-scale language-image (LLI) models
\cite{rombach2021highresolution,nichol2021glide,saharia2022photorealistic,ramesh2022hierarchical,yu2022scaling}. A novice user can gain significant control over the final image contents by combining text-based conditioning with unsupervised guidance from a reference image (usually a coarse stroke painting). The text prompt controls the overall image semantics, while the provided coarse stroke painting allows the user to define the color composition in the output scene.

Existing methods often aim attempt to achieve this through two means. The first category leverages conditional training using semantic segmentation maps \cite{rombach2021highresolution, gafni2022make,wang2022semantic}. However, the conditional training itself is quite time-consuming and requires a large scale collection of dense semantic segmentation labels across diverse data modalities. The second category, typically leverages an inversion based approach for mapping the input stroke painting to the target data manifold without requiring any paired annotations. For instance, a popular solution by \cite{meng2022sdedit, kim2022diffusionclip, song2020denoising} introduces the painting based generative prior by considering a noisy version of the original image as the start of the reverse diffusion process. However, the use of an inversion based approach causes an intrinsic domain shift problem if the domain gap between the provided stroke painting and the target domain is too high. In particular, we observe that the resulting outputs often lack details and resemble simplistic representations of the target domain.
For instance, in Fig.~\ref{fig:overview}, we notice that while the target domain consists of \emph{realistic photos} of a landscape, the generated outputs resemble simple pictorial arts which are not very realistic. Iteratively reperforming the guided synthesis with the generated outputs \cite{loopback} seems to improve realism but it is costly, 
some blurry details still persist (refer Fig.~\ref{fig:guided-synthesis}), 
and the generated outputs tend to lose faithfulness to the reference with each successive iteration. 

To address this,
we propose a 
diffusion-based 
guided image synthesis framework which models the output image as the solution of a constrained optimization problem 
(Sec.~\ref{sec:our-method}). Given a reference painting $y$, the constrained optimization is posed so as to find a solution $x$ with two constraints: 1) upon painting $x$ with an autonomous painting function we should recover a painting similar to reference $y$, and, 2) the output $x$ should lie in the target data subspace defined by the text prompt (\ie, if the prompt says \emph{``photo''} then we want the output images to be realistic photos instead of cartoon-like representations of the same concept). Subsequently, we show that while the computation of an exact solution for this optimization is infeasible, a practical approximation of the same can be achieved through simple gradient descent.

Finally, while the proposed optimization allows the user to generate image outputs with high realism and faithfulness (with reference $y$), the fine-grain semantics of different painting regions are inferred implicitly by the diffusion model. 
Such inference is typically dependent on the generative priors learned by the diffusion model, and might not accurately reflect the user's intent in drawing a particular region. For instance, in Fig.~\ref{fig:overview}, we see that the light blue regions can be inferred as blue-green grass instead of a river.
To address this, we show that by simply defining
a cross-attention based correspondence between the input
text tokens and user stroke-painting, the user
can control semantics of different painted regions without requiring semantic-segmentation based conditional training or finetuning. 

\section{Related Work}
\label{sec:related-work}

\textbf{GAN-based methods} have been extensively explored for performing guided image synthesis from coarse user scribbles. \cite{zhu2016generative,abdal2019image2stylegan,abdal2020image2stylegan++,karras2020training,richardson2021encoding,alaluf2021restyle,tov2021designing} use GAN-inversion for projecting user scribbles on to manifold of real images. While good for performing small scale inferences these methods fail to generate 
highly 
photorealistic outputs when the given stroke image is too far from the real image manifold.
Conditional GANs
\cite{park2019semantic,zhu2020sean,lee2020maskgan, esser2021taming, isola2017image,liu2019learning, sushko2020you} 
learn to directly generate realistic outputs based on user-editable semantic segmentation maps.
In another work, Singh \etal \cite{singh2022paint2pix} propose 
an image
synthesis framework which leverages autonomous painting agents \cite{singh2022intelli,singh2021combining,liu2021paint,zou2021stylized} for inferring photorealistic outputs from rudimentary user scribbles. 
Despite its efficacy, this requires the creation of a new dataset and conditional training for each target domain, which is expensive.

\textbf{Guided image synthesis with LLI models} \cite{rombach2021highresolution,nichol2021glide,saharia2022photorealistic,ramesh2022hierarchical,yu2022scaling,zhou2022towards,ding2021cogview}
has gained widespread attention \cite{gal2022image,liu2021more,kim2022diffusionclip,seo2022midms,kwon2022diffusion,huang2022draw,ruiz2022dreambooth} due to their ability to perform high quality image generation from diverse target modalities. 
Of particular interest are works wherein the guidance is provided using a coarse stroke painting and the model learns to generate outputs conditioned on both text and painting.
Current works in this direction typically 1) use semantic segmentation based conditional training \cite{rombach2021highresolution,gafni2022make,wang2022semantic} which is expensive, or, 2) adopt an inversion-based approach for mapping the input stroke painting to the target data manifold without requiring paired annotations. For instance,
Meng \etal \cite{meng2022sdedit,kim2022diffusionclip} propose 
guided image synthesis framework,
wherein the generative prior is introduced by simply considering a noisy version of the original sketch input as the start of the reverse diffusion process. 
Choi \etal \cite{choi2021ilvr} propose an iterative conditioning strategy wherein the intermediate diffusion outputs are successively refined to move towards the reference image. While effective, the use of an inversion-like approach causes an implicit domain shift problem, wherein the output images though faithful to the provided reference show blurry or less textured details. Iteratively reperforming
guided synthesis with generated outputs \cite{loopback} seems to
improve realism but it is costly. In contrast, we show that it is possible to perform highly photorealistic image synthesis while just requiring a single reverse diffusion pass. 

\begin{figure*}[h!]
\vskip -0.15in
\begin{center}
\centerline{\includegraphics[width=0.975\linewidth]{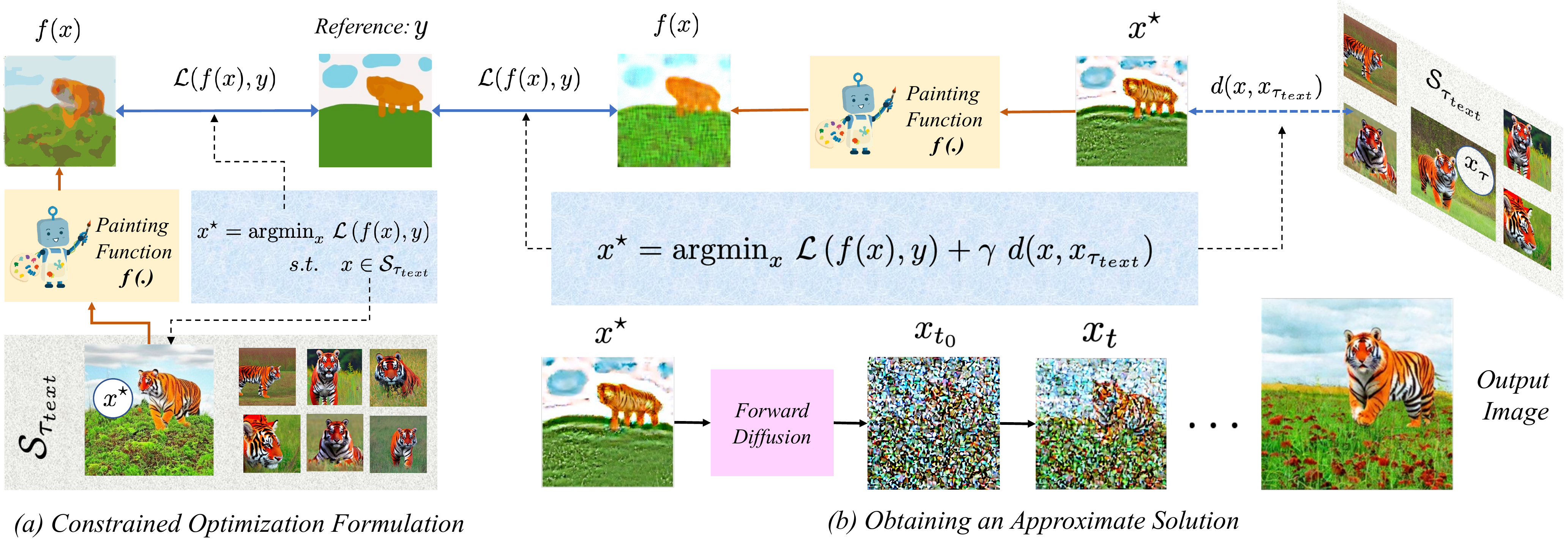}}
\caption{\emph{\textbf{Method Overview.}} (a) Given a reference painting $y$ and text prompt $\tau_{text}$, we first formulate the guided synthesis output as the solution $x^\star$ of a constrained optimization problem with 2 properties: 1) $x^\star$ lies in the subspace $\mathcal{S}_{\tau_{text}}$ of outputs conditioned only on the text, and, 2) upon painting $x$ we should recover reference painting $y$. While computing an exact solution of this optimization is infeasible, we show that an approximation can be obtained by solving the unconstrained optimization in (b). Here we first use gradient descent to compute a point $x^\star$ close to a random sample $x_{\tau_{text}} \in \mathcal{S}_{\tau_{text}}$, while still minimizing the painting loss $\mathcal{L}(f(x),y)$. This $x^\star$ is usually non-photorealistic due to gradient descent. We therefore use the diffusion based inversion from \cite{song2020denoising} to map it back to target domain $\mathcal{S}_{\tau_{text}}$.}
\label{fig:method-overview}
\end{center}
\vskip -0.2in
\end{figure*}

\textbf{Cross-attention control.} Recently, Hertz \emph{et al.} \cite{hertz2022prompt} propose a prompt-to-prompt image editing approach with text-conditioned diffusion models. By constraining the cross-attention features of all non-targeted text tokens to remain the same, they show that by only modifying the text prompt, it is possible to perform diverse image editing without changing the underlying structure of the original input image. In contrast, we use cross-attention control for from-scratch synthesis and show that by simply defining a cross-attention based correspondence between input text tokens and the user stroke-painting, it is possible to control and define the fine-grain semantics of different painted regions.

\section{Our Method}
\label{sec:our-method}

Let $f:\mathcal{D}_{real} \rightarrow \mathcal{D}_{paint}$ be a function mapping a real input image $x$ to its painted image $f(x)$.
Then given a colored stroke image $y$ and input text prompt $\tau_{text}$, we formulate the computation of output image $x^\star$ as the solution to the following  constrained optimization problem,
\begin{align}
    x^{\star} =\ &\text{argmin}_{x} \ \mathcal{L} \left(f(x),y\right) \label{eq:opt-1}\\ 
    &\text{subject to} \  x \in \mathcal{S}_{\tau_{text}}\label{eq:opt-2}
\end{align}
where $\mathcal{L} \left(f(x),y\right)$ represents a distance measure between the painted output $f(x)$ of image $x$ and the target painting $y$, while $\mathcal{S}_{\tau_{text}}$ represents the subspace of output images conditioned only on the text input.

In other words, by additionally conditioning on a stroke image $y$, we wish to find a solution $x^\star$ such that 1) the distance between the painted image of $x$ and reference painting $y$ is minimized, while at the same time ensuring 2) the final solution lies in the subspace of images conditioned only on the text prompt $\tau_{text}$. For instance, if the text says \emph{``a realistic photo of a tree''} then the use of stroke-based guidance should still produce a \emph{``realistic photo''},  wherein the composition of the tree regions is controlled by the painting $y$.

\subsection{GradOP: Obtaining an Approximate Solution}
\label{sec:bilevel}

The optimization problem in Eq.~\ref{eq:opt-1} can be reformulated as an unconstrained optimization problem as,
\begin{align}
    x^{\star} = \text{argmin}_{x} \ \mathcal{L} \left(f(x),y\right) + \gamma \ d(x,\mathcal{S}_{\tau_{text}}), \label{eq:opt-3}
\end{align}
where $d(x,\mathcal{S}_{\tau_{text}})$ represents a distance measure between $x$ and subspace $\mathcal{S}_{\tau_{text}}$, and $\gamma$ is a hyperparameter.

A cursory glance at the above formulation should make it evident that the computation of an exact solution is infeasible without first generating a large enough sample size for the $\mathcal{S}_{\tau_{text}}$ subspace, which will be quite time consuming. 

To address this, we propose to obtain an approximate solution by estimating
$d(x,\mathcal{S}_{\tau_{text}})$ through the distance of $x$ from a single random sample $x_{\tau_{text}} \in \mathcal{S}_{\tau_{text}}$. Thus, we can approximate the optimization problem as follows,
\begin{align}
    x^{\star} = \text{argmin}_{x} \ \mathcal{L} \left(f(x),y\right) + \gamma \ d( x , x_{\tau_{text}}). \label{eq:opt-4}
\end{align}
Assuming a latent diffusion model with decoder $\mathcal{D}$, we can rewrite the above above optimization in latent space as,
\begin{align}
    z^\star = \text{argmin}_{z} \ \mathcal{L} \left(f(\mathcal{D}(z)),y\right) + \gamma \ \Vert z - z_{\tau_{text}}\Vert_2.  \label{eq:opt-5}
\end{align}
where the image output $x^\star$ can be computed as $x^\star = \mathcal{D}(z^\star)$.

In order to solve the above optimization problem, we first use the diffusion model to sample  $x_{\tau_{text}} \in \mathcal{S}_{\tau_{text}}$. Initializing $z=\mathcal{E}(x_{\tau_{text}})$, where $\mathcal{E}$ represents the encoder, we solve the above optimization using gradient descent (assuming $f$ and $\mathcal{L}$ are differentiable). Finally, we note that the solution $x^{\star}=\mathcal{D}(z^{\star})$ to the above approximation of the optimization problem might be non-photorealistic due to gradient descent. We therefore use the diffusion based inversion approach from \cite{kim2022diffusionclip} in order to map it to the target image manifold. Please refer Alg.~\ref{alg:bilevel} for the detailed implementation.

\begingroup
\begin{algorithm}[t]
	\caption{GradOP: Solution Approximation}
    \textbf{Input}: Stroke Painting $y$, text prompt $\tau_{text}$\\
    \textbf{Output}: Output image $x$ conditioned on both $\tau_{text}, y$\\
    \textbf{Require}: Differentiable painting function $f$, differentiable distance measure $\mathcal{L}$, hyperparameter $\gamma, t_0$.
    \begin{algorithmic}[1]
            \State Sample \ $x_{\tau_{text}} \in \mathcal{S}_{\tau_{text}}$;
            \State Initialize \ $z = z_{\tau_{text}} = \mathcal{E}(x_{\tau_{text}})$;
            \For{$0 \leq i \leq M$}
                \State $\mathcal{L}_{total} = \mathcal{L} \left(f(\mathcal{D}(z)),y\right) + \gamma \ \Vert z - z_{{\tau}_{text}}\Vert_2$;
                \State $z = z - \lambda \nabla_z \mathcal{L}_{total}$;
            \EndFor
            \State $ z_{t_0} = \textsc{ForwardDiff}(z^\star=z,0\rightarrow t_0)$;
            \State $z = \textsc{ReverseDiff}(z_{t_0},t_0 \rightarrow 0)$;
            \State \Return $x_{out} = \mathcal{D}(z)$.
	\end{algorithmic}
	\label{alg:bilevel}
\end{algorithm}
\endgroup

\subsection{GradOP+ : Improving Sampling Efficiency}
\label{sec:igrad}

\begin{figure}[h!]
\begin{center}
\centerline{\includegraphics[width=1.\linewidth]{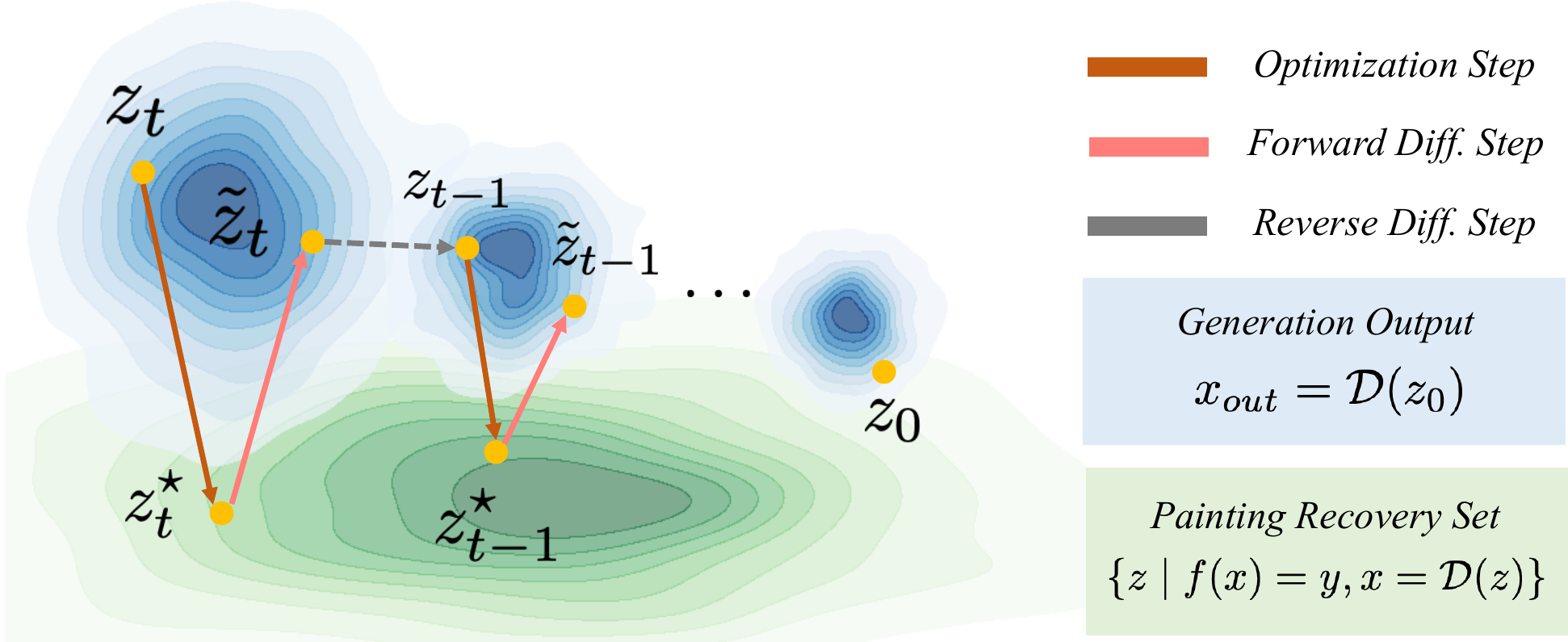}}
\vskip -0.05in
\caption{
\emph{\textbf{GradOP+ Overview.}} 
At any timestep $t$, the optimization in Eq.~\ref{eq:opt-6} ($z_t \rightarrow z^\star_t$) reduces the painting recovery loss, while the forward diffusion step $z^\star_t\rightarrow \Tilde{z}_t$ maps it back to the expected latent distribution. By iteratively performing this optimization, \emph{GradOP+} modifies the reverse sampling trajectory to lead to output $x_{out} = \mathcal{D}(z_0)$ which is also faithful to the target painting $y$.
}
\label{fig:gradop+}
\end{center}
\vskip -0.2in
\end{figure}

While the guided synthesis solution in Alg.~\ref{alg:bilevel} shows great output results,
for each output image it first requires the sampling of a text-only conditioned output $x_{\tau_{text}} \in \mathcal{S}_{\tau_{text}}$. 
To address this, we propose a modified guided image synthesis approach which allows for equally high quality outputs while requiring just a single reverse diffusion pass 
for each output. 
Our key insight is that a lot of information in $z^\star$ is discarded during the forward diffusion from $z^\star \rightarrow z_{t_0}$. Thus, instead of performing the optimization to first compute $z^\star$, we would like to directly 
optimize the intermediate latent states $z_t$ by injecting the optimization gradients within the reverse diffusion process itself (refer Fig.~\ref{fig:gradop+}).

In particular, at any timestep $t$ during the reverse diffusion, we wish to introduce optimization gradients in order to solve the following optimization problem,
\begin{align}
    z^{\star}_t = \text{argmin}_{z} \ \mathcal{L} \left(f(\mathcal{D}(z)),y\right) + \gamma \ \Vert z - z_{t}\Vert_2. \label{eq:opt-6}
\end{align}
However, the introduction of gradients will cause $z^\star_t$ to not conform with the expected latent distribution at timestep $t$. We therefore pass it through the forward diffusion process in order to map it back to the expected latent variable distribution. 
Please refer Alg.~\ref{alg:igrad} for the detailed implementation. 

\begin{algorithm}[tbh]
	\caption{GradOP+ : Improving Sampling Efficiency}
    \textbf{Input}: Stroke Painting $y$, text prompt $\tau_{text}$\\
    \textbf{Output}: Output image $x$ conditioned on both $\tau_{text}, y$\\
    \textbf{Require}: Differentiable painting function $f$, distance measure $\mathcal{L}$, hyperparameter $\gamma, t_0, t_{start}, t_{end}$.
    \begin{algorithmic}[1]
            \State Sample \ $z_{T} \sim \mathcal{N}(0,\bm{I})$;
            \For{$0 \leq t < T$}
                \State $z_t = \textsc{ReverseDiff}(z_{t+1},t+1 \rightarrow t)$;
                \If{$t_{start} \leq t \leq t_{end}$}
                    \State Initialize \ $z = z_{t}$;
                    \For{$0 \leq i \leq M$}
                        \State $\mathcal{L}_{total} = \mathcal{L} \left(f(\mathcal{D}(z)),y\right) + \gamma \ \Vert z-z_{t}\Vert_2$;
                        \State $z = z - \lambda \nabla_z \mathcal{L}_{total}$;
                    \EndFor
                    \State $z_t = \textsc{ForwardDiff}(z^
                    \star_t=z,0 \rightarrow t)$
                \EndIf
            \EndFor
            \State \Return $x_{out} = \mathcal{D}(z_0)$.
	\end{algorithmic}
	\label{alg:igrad}
\end{algorithm}

\subsection{Controlling Semantics of Painted Regions}
\label{sec:semantic-injection}

Finally, while the above approximate guided image synthesis algorithm allows for generation of image outputs with high \emph{faithfulness} and \emph{realism}, the semantics of different painted regions are inferred in an implicit manner.    
Such inference is typically based on the cross-attention priors (learned by the diffusion model) between the provided text tokens and the input painting throughout the reverse diffusion process. For instance, in the first example from Fig.~\ref{fig:semantic-injection}, we note that for different outputs, the blue region can be inferred as a river, waterfall, or a valley. Also note that some painting regions might be entirely omitted (\eg the brown strokes for the hut), if the model does not understand that the corresponding strokes indicate a distinct semantic entity \eg a hut, small castle \etc. Moreover, as shown in Fig.~\ref{fig:semantic-injection} such discrepancies persist even if the corresponding text tokens (\eg a hut) are added to the textual prompt. 

Our key motivation is that the when generated faithfully, the average attention maps across different cross-attention layers show high overlap with the target object segmentation during the initial to intermediate parts of the reverse diffusion process. In our experiments, we found the reverse to also be true. That is, by constraining the cross attention map corresponding to a target semantic label to have a high overlap with the desired painting region, it is possible to control the semantics of different painting regions without the need for segmentation based conditional training.

In particular, given the binary masks corresponding to different painting regions $\{\mathcal{B}_1, \dots \mathcal{B}_N\}$ and the corresponding semantic labels $\{u_1, \dots u_N\}$, we first modify the input text tokens as follows,
\begin{align}
    \tau_{modified} = \tau + \left\{ \textsc{CLIP}(u_i) \mid i \in [1,N] \right\},
\end{align}
where $\tau$ is the set of CLIP \cite{radford2021learning} tokens for input text prompt. 

At any timestep $t \in [0,T]$ during the reverse diffusion process, we then enforce semantic control by modifying the cross-attention map $\mathcal{A}^i_t$ corresponding to label $u_i$ as follows,
\begin{align}
    \Tilde{\mathcal{A}^i_t} = w_i \left[(1-\kappa_t) \ \mathcal{A}^i_t + \kappa_t \ \frac{\mathcal{B}_i \ }{\Vert \mathcal{B}_i \Vert_F} \ \Vert \mathcal{A}_t^i \Vert_F \right]
\end{align}
where $\Vert.\Vert_F$ represents the Frobenius norm, $\kappa_t = t/T \in [0,1]$ helps regulate the overlap between the cross-attention output $\Tilde{\mathcal{A}^i_t}$ and the desired painting region $\mathcal{B}_i$ during the reverse diffusion process, 
and, weights $w_i, \ i\in[1,N]$ help the user to control the relative importance of expressing different semantic concepts in the final image.


\begin{figure*}[h!]
\vskip -0.1in
\begin{center}
\centering
     \begin{subfigure}[b]{1.\textwidth}
         \centering
         \includegraphics[width=\textwidth]{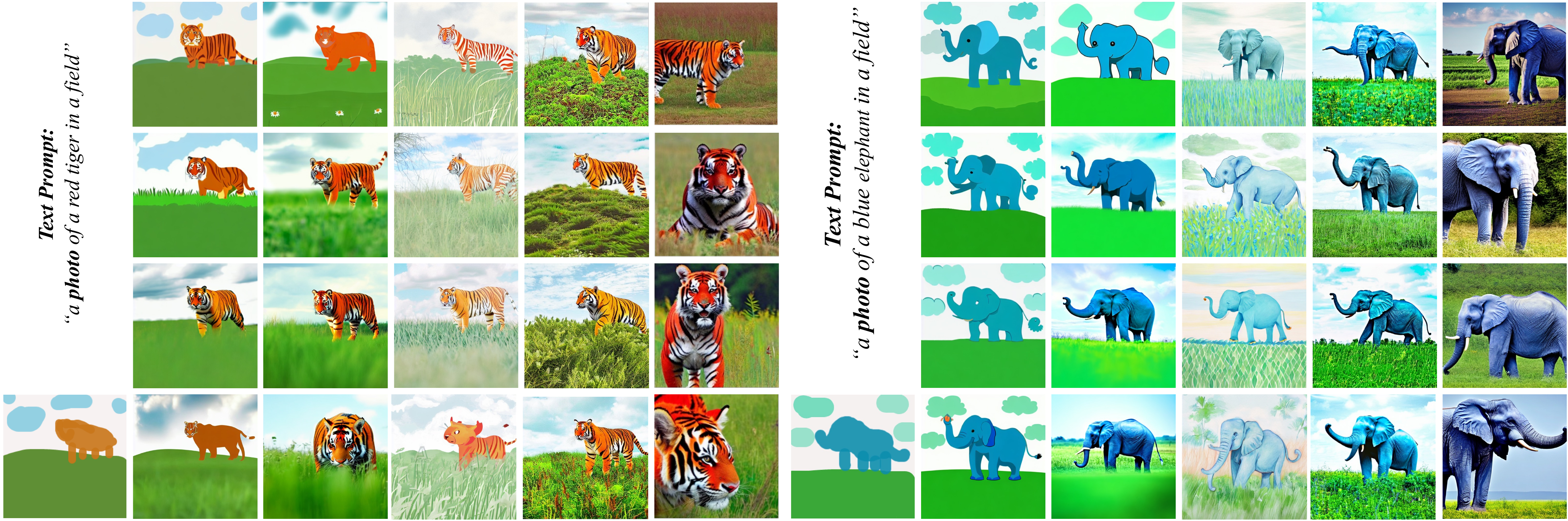}
     \end{subfigure} 
     \vskip 0.02in
     \hfill
      \begin{subfigure}[b]{1.\textwidth}
         \centering
         \includegraphics[width=\textwidth]{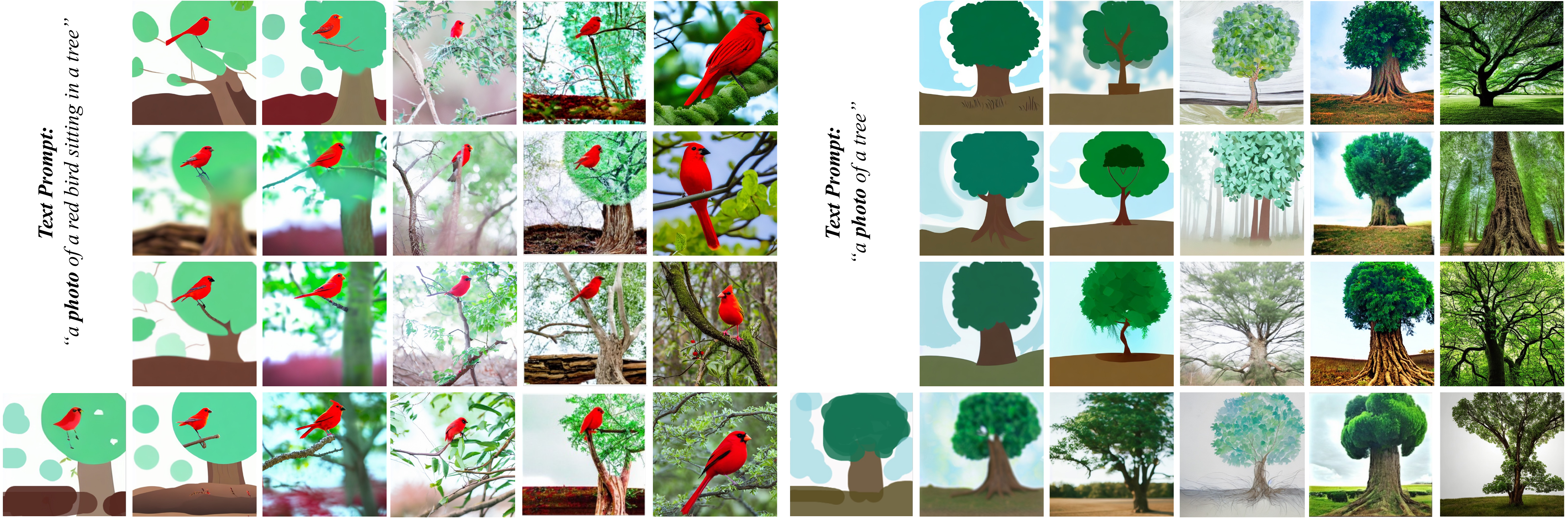}
     \end{subfigure}
     \vskip 0.02in
     \hfill
      \begin{subfigure}[b]{1.\textwidth}
         \centering
         \includegraphics[width=\textwidth]{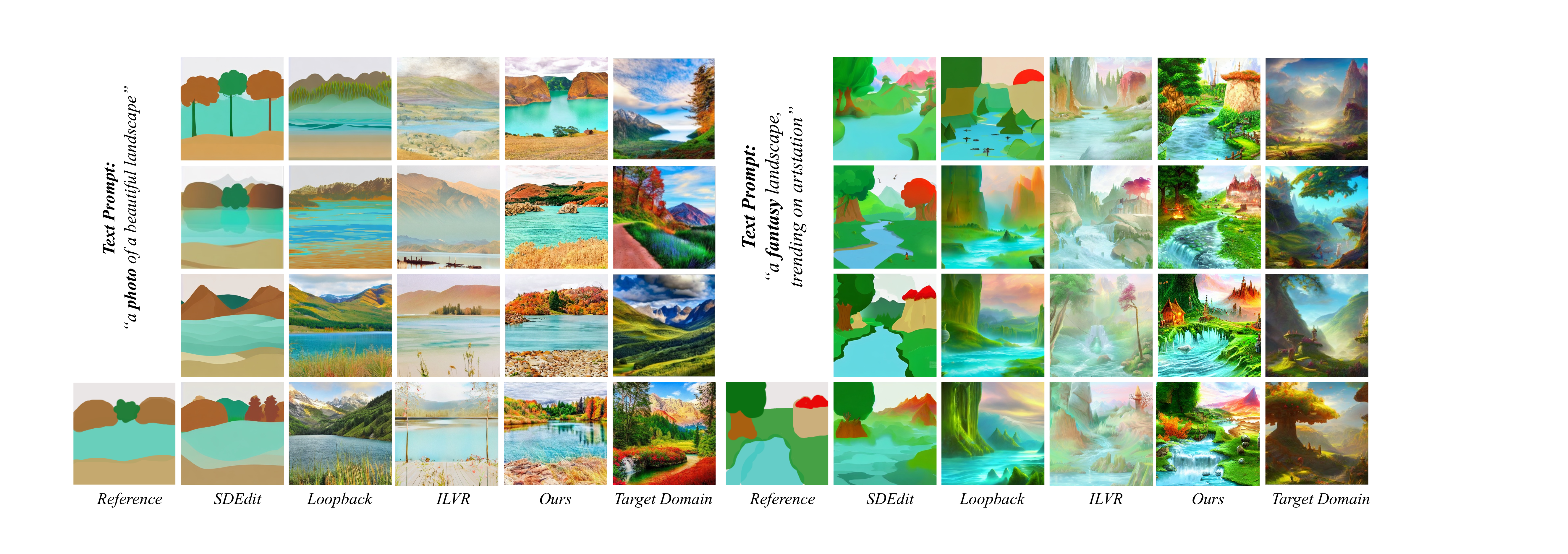}
     \end{subfigure}
\vskip -0.05in
\caption{\emph{\textbf{Qualitative comparisons.}} We compare the performance of our approach with prior works \cite{meng2022sdedit,loopback,choi2021ilvr} based on their \emph{faithfulness} to the provided reference, and the \emph{realism} with respect to the target domain (generated by conditioning only on the text prompt). Please note that for our results, we show the \emph{GradOP} (Alg.~\ref{alg:bilevel}) and \emph{GradOP+} (Alg.~\ref{alg:igrad}) outputs in row 1,2 and row-3,4 respectively. 
}
\label{fig:guided-synthesis}
\end{center}
\vskip -0.3in
\end{figure*}

\section{Experiments}
\label{sec:experiments}

\emph{\textbf{Implementation Details.}}  We use publicly available text-conditioned latent diffusion models \cite{diffusers, rombach2021highresolution} for implementing the purposed approach in Sec.~\ref{sec:our-method}. The constrained optimization is performed using gradient descent with the Adam \cite{kingma2014adam} optimizer and number of gradient steps $N_{grad}\in[20,60]$ (please refer Sec.~\ref{sec:var-ngrad} for detailed analysis). While several formulations of the distance measure $\mathcal{L}$ and painting function $f$ are possible (refer supp. material for details), we find that simply approximating the function $\mathcal{L}$ using mean squared distance and $f$ as a convolution operation with a gaussian kernel seems to give the fastest inference time performance with our method. For consistency reasons, we use the non-differentiable painting function from SDEdit \cite{meng2022sdedit} while reporting quantitative results
(refer Sec.~\ref{sec:stroke-based-synthesis}). 

\subsection{Stroke Guided Image Synthesis}
\label{sec:stroke-based-synthesis}
\emph{\textbf{Evaluation metrics.}}
Given an input stroke painting, we compare the performance of our approach with prior works in guided image synthesis when no paired data is available. The performance of the final outputs is measured in terms of both \emph{faithfulness} of the generated image with the target stroke painting as well as the \emph{realism} of the final output distribution. In particular, given an input painting $y$ and output real image prediction $x$, we define faithfulness $\mathcal{F}(x,y)$ as,
\begin{align}
    \mathcal{F}(x,y) = \mathcal{L}_2(f(x),y)
\end{align}
where $f(.)$ is the painting function. Thus an output image $x$ is said to have high faithfulness with the given painting $y$ if upon painting the final output $x$ we get a painting $\Tilde{y}=f(x)$ which is similar to the original target painting $y$. 

Similarly, given a set of output data samples $\mathcal{S}({y,\tau_{text}})$ conditioned on both painting $y$ and text $\tau_{text}$, and, $\mathcal{S}({\tau_{text}})$ conditioned only on the text, the \emph{realism} $\mathcal{R}$ is defined as,
\begin{align}
    \mathcal{R}(\mathcal{S}({y,\tau_{text}})) = FID \left(\mathcal{S}({y,\tau_{text}}),\mathcal{S}({\tau_{text}})\right)
\end{align}
where $FID$ represents the Fisher inception distance \cite{heusel2017gans}. 

\emph{\textbf{Baselines.}} We compare our approach with prior works on guided image synthesis from stroke paintings with no paired data. In particular we show comparisons with, \emph{1) SDEdit} \cite{meng2022sdedit} wherein the generative prior is introduced by first passing the painting $y$ through the forward diffusion pass $y \rightarrow y_{t_0}$ \cite{kim2022diffusionclip,song2020denoising}, and then performing reverse diffusion $y_{t_0} \rightarrow y_0$ to get the output image $x = y_0$\footnote{Please note that due to space constraints, we primarily use the standard hyperparameter value of $t_0 = 0.8$ in the main paper, and refer the reader to the supp. material for detailed comparisons with changing $t_0 \in [0,1]$.}.
\emph{2)  SDEdit + Loopback} \cite{loopback} which reuses the last diffusion output to iteratively increase the realism of the final output. \emph{3) ILVR}\footnote{Please note that the original ILVR \cite{choi2021ilvr} algorithm was proposed for iterative refinement with diffusion models in pixel space. We adapt the ILVR implementation for inference with latent diffusion models \cite{rombach2021highresolution} for the purposes of this paper. Please refer supp. material for further details.}\cite{meng2022sdedit}: which uses an iterative refinement approach for conditioning the output $x$ of the diffusion model with a guidance image $y$. Unless otherwise specified, we use the GradOP+ algorithm 
(refer Alg.~\ref{alg:igrad}) 
when reporting evaluation results.

\begingroup
\setlength{\tabcolsep}{5.0pt}
\small
\begin{table}[t]
\begin{center}
\small
\begin{tabular}{l|cc|cc}
\toprule
\multirow{2}{*}{Method} & \multicolumn{2}{c|}{Evaluation criteria} & \multicolumn{2}{c}{User Study Results} \\
\cline{2-5} 
 & $\mathcal{F}({x,y})\downarrow$ & $\mathcal{R}(.)\downarrow$ & Realism $\uparrow$ & Satisfaction $\uparrow$ \\
\hline
SDEdit \cite{meng2022sdedit} & 88.93 & 223.8  & 94.09 \% &  91.98\% \\ 
Loopback \cite{loopback} & 104.6 & 132.9  & 54.28 \% & 85.32\% \\
ILVR \cite{choi2021ilvr} & 108.2 & 161.7  & 76.54 \% & 93.47\% \\
\textbf{Ours} & 94.40 & 134.2  & \text{N/A} & \text{N/A} \\ %
\bottomrule
\end{tabular}
\end{center}
\vskip -0.2in
\caption{\emph{\textbf{Quantitative Evaluations}}. \emph{(Left)} Method comparison w.r.t \emph{faithfulness} $\mathcal{F}$ to the reference painting and \emph{realism} $\mathcal{R}$ to the target domain. \emph{(Right)} User-study results, showing \% of inputs for which human subjects prefer our approach over prior works.}
\label{tab:quant-results}
\vskip -0.15in
\end{table}
\endgroup

\begin{figure*}[h!]
\vskip -0.1in
\begin{center}
\centerline{\includegraphics[width=0.95\linewidth]{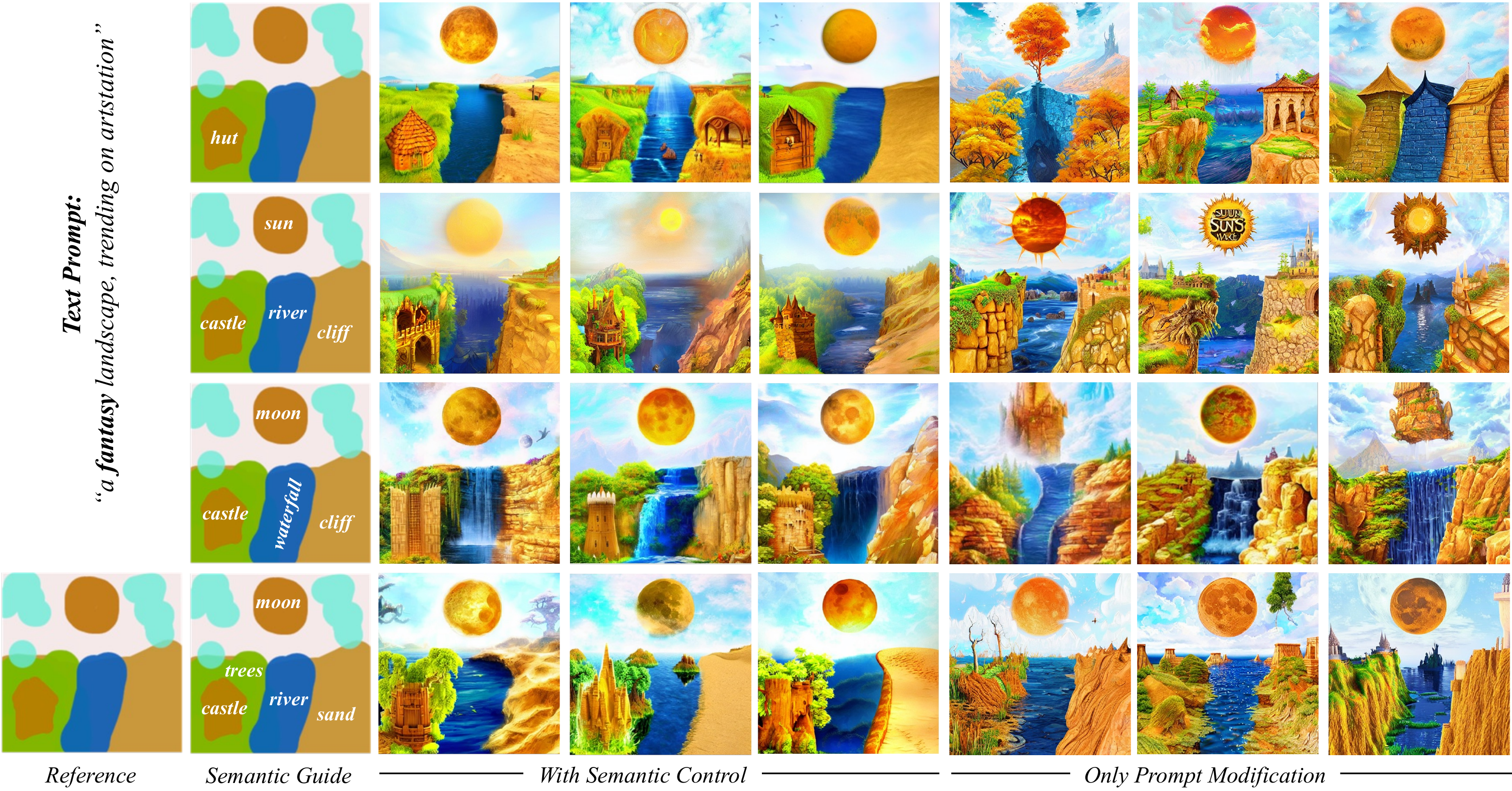}}
\vskip -0.05in
\caption{\emph{\textbf{Controlling semantics of different painted regions.}} We compare image generation outputs (Col 3-5) using the cross-attention control approach from Sec.~\ref{sec:semantic-injection} with outputs (Col 6-8) generated by only modifying the input text prompt. Note that for each semantic guide (Col 2), the text prompt modification is performed by adding the corresponding semantic labels at the end of the text prompt. For instance, the modified text prompt for examples in row-1 would be ``a fantasy landscape, trending on artstation showing a hut''.}
\label{fig:semantic-injection}
\end{center}
\vskip -0.35in
\end{figure*}

\emph{\textbf{Qualitative Results.}} Results are shown in Fig.~\ref{fig:guided-synthesis}. We observe that both proposed approximate optimization methods (\ie \emph{GradOP} in row-1,2 and \emph{GradOP+} in row-3,4) lead to output images which are both highly photorealistic as well as \emph{faithful} with reference painting. 
In contrast, while 
SDEdit \cite{meng2022sdedit} shows high faithfulness to the input painting, the final outputs lack details and resemble more of a pictorial art rather than realistic photos. 
Iteratively reperforming the guided synthesis with the generated outputs (SDEdit + Loopback \cite{loopback}) helps improve the realism of output images,
however, we find that this has two main disadvantages. First, the iterative loop increases the effective time required for generating each data sample (\eg four reverse sampling steps instead of just one). Second, we note that as the number of successive iterations increase the final outputs become less and less faithful to the original painting input. Finally, ILVR \cite{choi2021ilvr} leads to more realistic outputs, however, the final outputs are not fully faithful to the reference painting in terms of the overall color composition.

\emph{\textbf{Quantitative Results.}}
In addition to qualitative results we also quantitatively evaluate the final outputs on the \emph{faithfulness} $\mathcal{F}(x,y)$ and \emph{targeted-realism} $\mathcal{R}(.)$ metrics defined earlier. Additionally, similar to \cite{meng2022sdedit} we also perform a human user study wherein the \emph{realism} and the overall satisfaction score (\emph{faithfulness} + \emph{realism}) are evaluated by human subjects (please refer supp. material for details). Results are shown in Tab.~\ref{tab:quant-results}. We find that as expected, while SDEdit \cite{meng2022sdedit} leads to the best faithfulness with the target painting, it exhibits very poor performance in terms of the \emph{realism} score. SDEdit with loopback \cite{loopback} improves the realism score but the resulting images start loosing faithfulness with the given reference. In contrast, our approach leads to the best tradeoff between faithfulness to the target image and realism with respect to the target domain. 
These findings are also reflected in the user-study results wherein our method is preferred by $>85.32\%$ of human subjects in terms of the overall satisfaction score.

\subsection{Controlling Semantics of Painted Regions}

Results are shown in Fig.~\ref{fig:semantic-injection}. We observe that in absence of semantic attention control, the model tries to infer the semantics of different painting regions in an implicit manner. For instance, the orange strokes in the sky region can be inferred as the sun, moon, or even as a yellow tree. Similarly, the brown strokes in the lower-left region (intended to draw a \emph{hut} or small \emph{castle}) are often inferred as muddy or rocky parts of the terrain. Moreover, such disparity continues even after modifying the input prompt to describe the intended semantic labels. For instance, in row-1 from Fig.~\ref{fig:semantic-injection}, while changing the text prompt to include the text ``hut'' leads to the emergence of ``hut'' like structures, the inference is often done in a manner that is not intended by the user. 

In contrast, by ensuring a high overlap between the intended painting regions and the cross-attention maps for the corresponding semantic labels (refer Sec.~\ref{sec:semantic-injection}), we are able to generate outputs which follow the intended semantic guide in a much more accurate manner. For instance, the user is able to explicitly specify that the brown regions on the ground describes a hut (row 1) or castle (row 2-4). Similarly, the semantics of different regions can be controlled, \eg the blue region is specified as a river or waterfall, the orange strokes in the sky is specified as moon or sun \etc.

\section{Analysis}
\label{sec:analysis}

\subsection{Variation in Target Domain}

\begin{figure*}[htbp]
\begin{center}
\centering
     \begin{subfigure}[b]{0.48\textwidth}
         \centering
         \includegraphics[width=\textwidth]{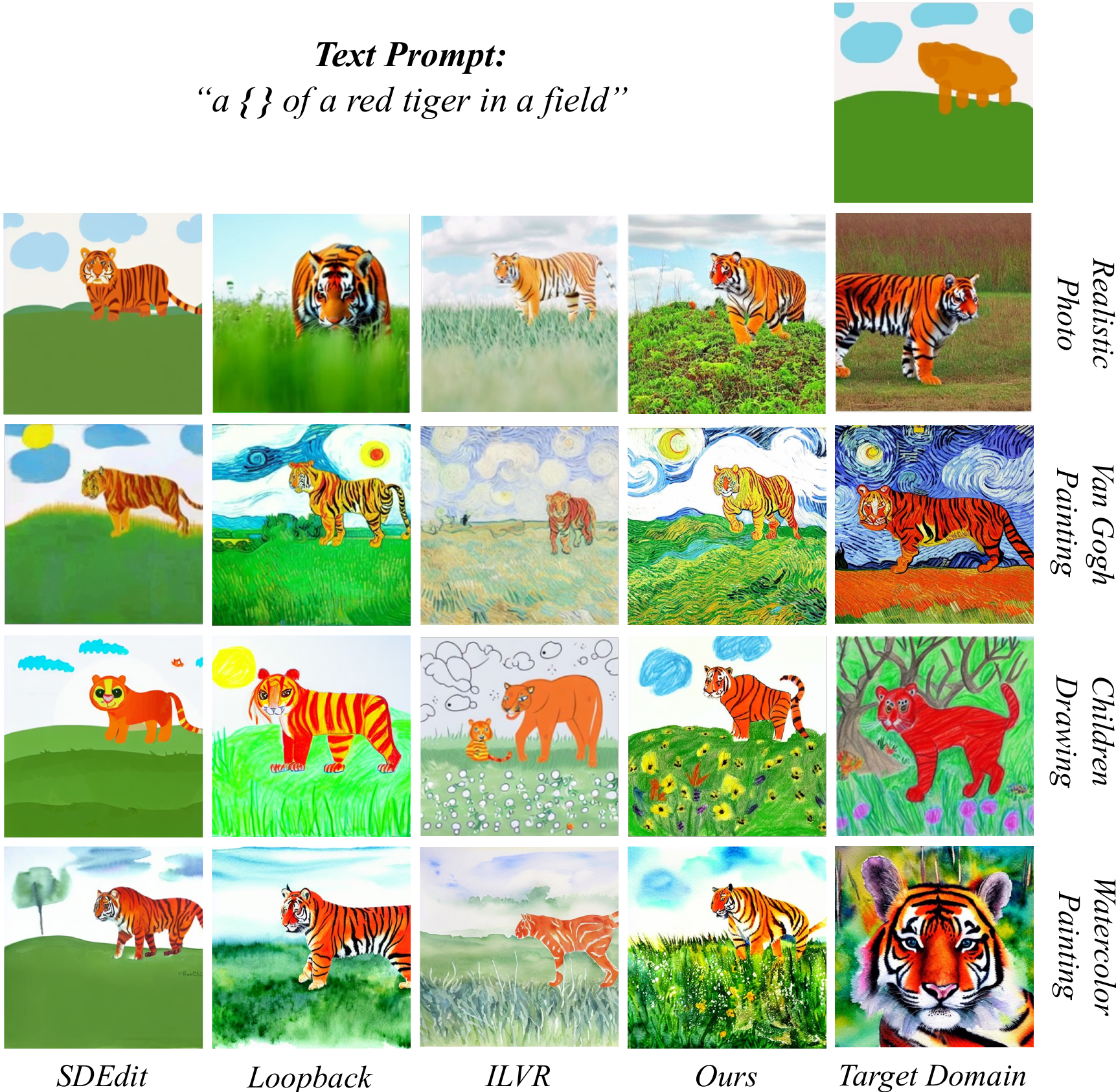}
     \end{subfigure}
     \hfill
      \begin{subfigure}[b]{0.48\textwidth}
         \centering
         \includegraphics[width=\textwidth]{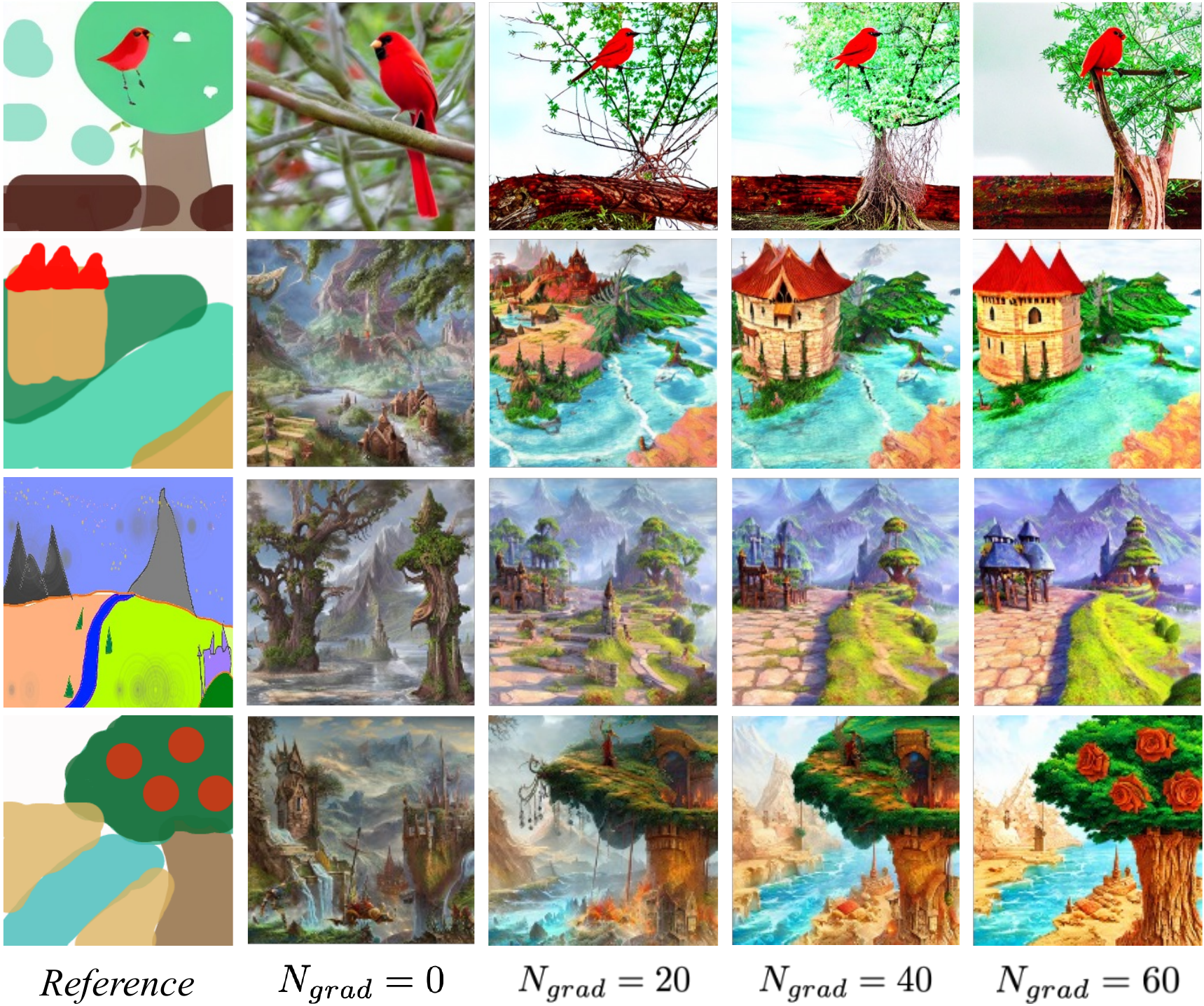}
     \end{subfigure} 
\vskip -0.05in
\caption{\emph{\textbf{Method Analysis.}} Comparing guided image synthesis performance across \emph{(left)} variation in target domain, and \emph{(right)} variation in number of gradient descent steps $N_{grad}$ used for performing the proposed optimization. Please zoom-in for best comparisons.}
\label{fig:var-domain}
\end{center}
\vskip -0.2in
\end{figure*}

In this section, we analyse the generalizability of the our approach across different target domains (\eg children drawings, disney scenes) and compare the output performance with prior works. Results are shown in Fig.~\ref{fig:var-domain}-a. We observe that our approach is able to adapt the final image outputs reliably across a range of target domains while still maintaining a high level of faithfulness with the target image.
In contrast, SDEdit \cite{meng2022sdedit} generates outputs which lack details and thereby look very similar across a range of target domains.
SDEdit + Loopback \cite{loopback} addresses this problem to some extent, but it requires multiple reverse diffusion passes and the generated outputs tend to lose their faithfulness to the provided reference with each iteration.

\begin{figure}[t]
\begin{center}
\centering
     \begin{subfigure}[b]{1\columnwidth}
         \centering
         \includegraphics[width=\textwidth]{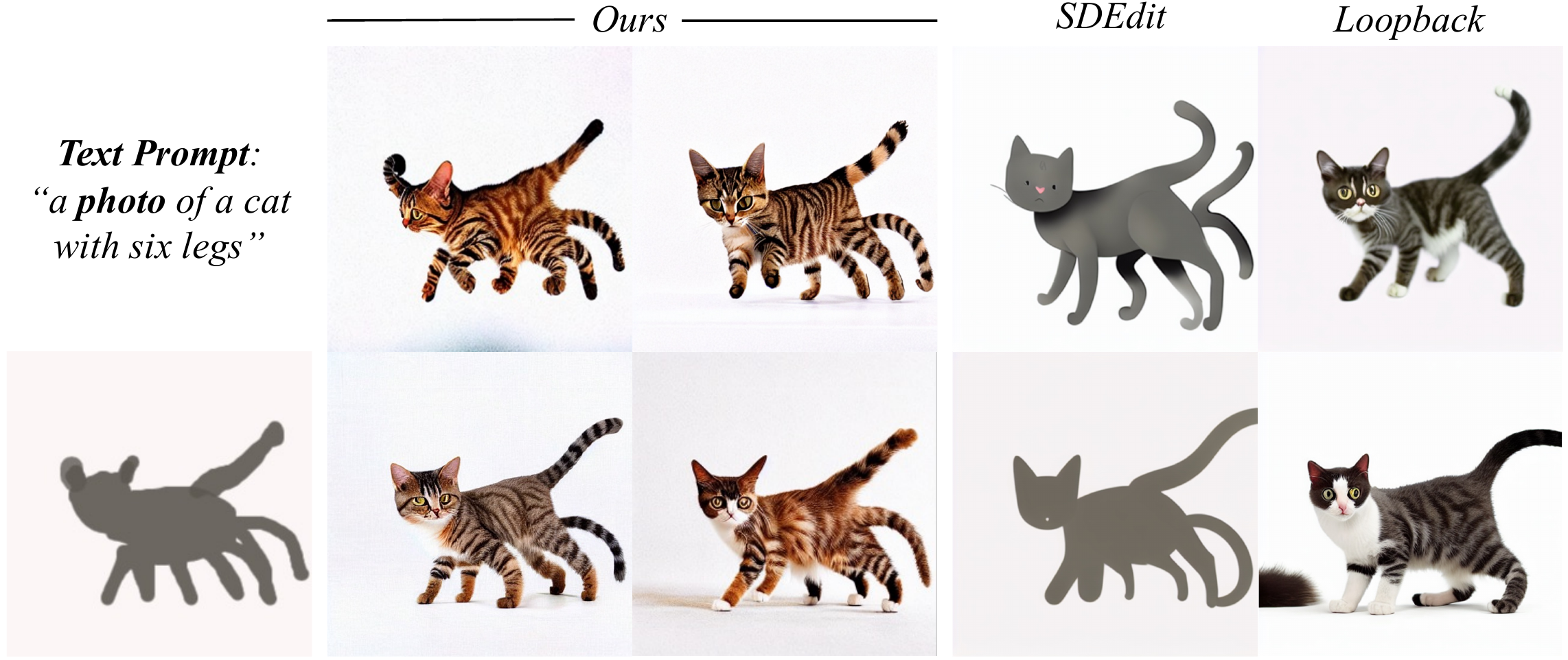}
     \end{subfigure}
     \hfill
      \begin{subfigure}[b]{1\columnwidth}
         \centering
         \includegraphics[width=\textwidth]{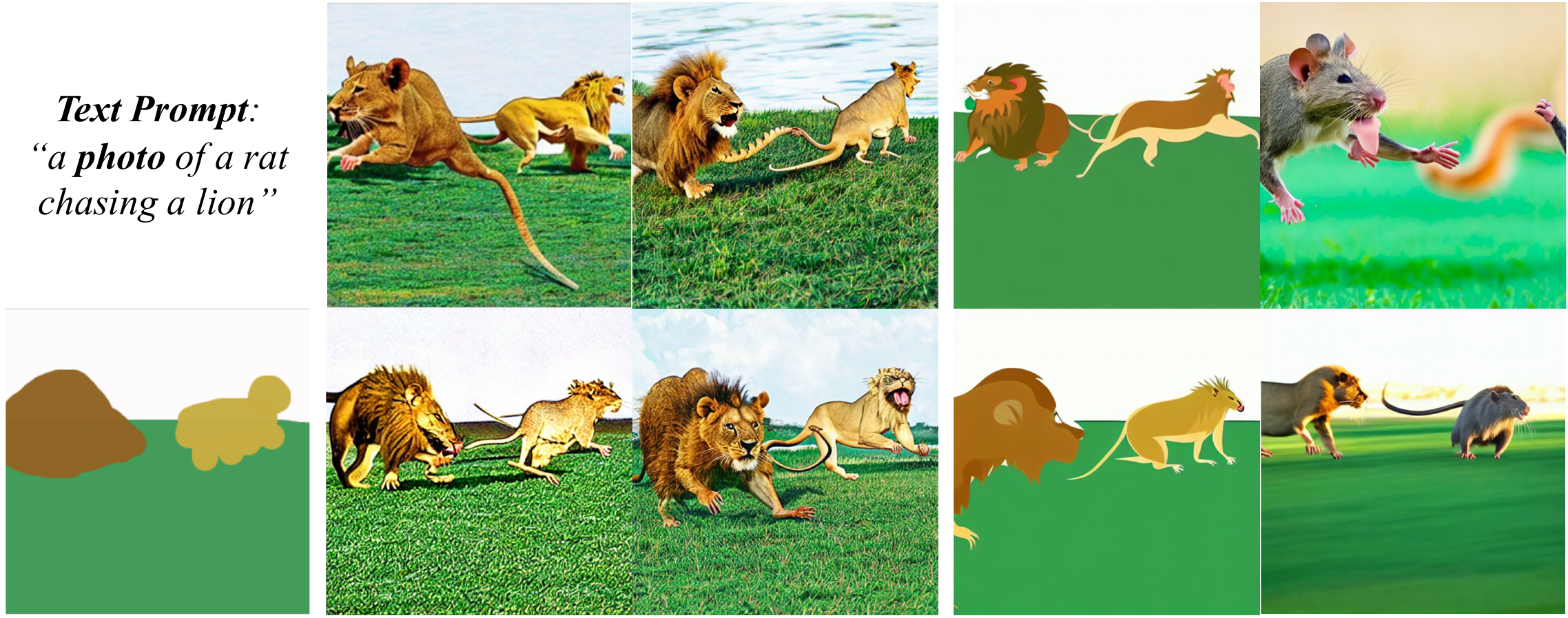}
     \end{subfigure} 
\vskip -0.05in
\caption{\textbf{\emph{Out-of-distribution performance.}} Analysing \emph{success} (top) and \emph{failure} (bottom) cases for out-of-distribution prompts.}
\label{fig:out-of-dist}
\end{center}
\vskip -0.3in
\end{figure}

\subsection{Variation with Number of Gradient Steps}
\label{sec:var-ngrad}
In this section, we analyse the variation in output performance as we change the number of gradient descent steps $N_{grad}$ used to solve the unconstrained optimization problem in Sec.~\ref{sec:our-method}. Results are shown in Fig.~\ref{fig:var-domain}-b. As expected, we find that for $N_{grad}=0$, the generated outputs are sampled randomly from the subspace of outputs $(\mathcal{S}_{\tau_{text}})$ conditioned only on the text. As the number of gradient-descent steps increase, the model converges to a subset of solutions within the target subsapce $\mathcal{S}_{\tau_{text}}$ which exhibit higher \emph{faithfulness} with the provided reference. Please note that this behaviour is in contrast with SDEdit \cite{meng2022sdedit}, wherein the increase in \emph{faithfulness} to the reference is corresponded with a decrease in the \emph{realism} of the generated outputs \cite{meng2022sdedit}.

\subsection{Out-of-Distribution Generalization}
\label{sec:out-of-dist}

As shown in Sec.~\ref{sec:experiments}, \ref{sec:analysis}, we find that the proposed approach allows for a high level of semantic control (both color composition and fine-grain semantics) over the output image attributes, while still maintaining the \emph{realism} with respect to the target domain. Thus a natural question arises: \emph{Can we use the proposed approach to generate realistic photos with out-of-distribution text prompts?} 

As shown in Fig.~\ref{fig:out-of-dist}, we observe that both success and failure cases exist for out-of-distribution prompts. For instance, while the model was able to generate ``\emph{realistic} photos of cats with six legs'' 
(note that for the same inputs prior works either 
generate faithful but cartoon-like outputs, or, simply generate regular cats), 
it shows poor performance while generating ``a photo of a rat chasing a lion''.

\section{Conclusions}
\label{sec:conclusion}

In this paper, we present a novel framework for performing guided image synthesis synthesis with user scribbles, without the need for paired annotation data. 
We point that prior works in this direction \cite{meng2022sdedit,loopback,choi2021ilvr}, typically adopt an inversion-like approach which leads to outputs which lack details and are often simplistic representations of the target domain. To address this, we propose a novel formulation which models the guided synthesis output as the solution of a constrained optimization problem. While obtaining an exact solution to this optimization is infeasible, we propose two methods \emph{GradOP} and \emph{GradOP+} which try to obtain an approximate solution to the constrained optimization in a sample-efficient manner. Additionally, we show that by defining a cross-attention based correspondence between the input text tokens and user painting,  it is possible to control semantics of different painted regions without the need for semantic segmentation based conditional training.

\newpage
{\small
\bibliographystyle{ieee_fullname}
\bibliography{gradop}
}

\newpage
\appendix

\title{Supplementary Material \\ 
High-Fidelity Guided Image Synthesis with Latent Diffusion Models
}

\twocolumn[{
\maketitles
}]

\section{Additional Results}

In this section, we provide additional results which could not be included in the main paper due to space constraints. In particular, we note that baseline methods like SDEdit \cite{meng2022sdedit} can often be run using different values of the hyperparameter $t_0$. We therefore provide additional results comparing the performance of SDEdit at different $t_0 \in [0,1]$ (refer Sec.~\ref{sec:sdedit-var}). Additionally, we introduce some custom baselines (which could be used for improving the realism of final image outputs) and show results comparing their output performance with our approach (refer Sec.~\ref{sec:custom-baselines}).

\subsection{Additional Comparisons with SDEdit}
\label{sec:sdedit-var}

Recall, given a stroke painting $y$, SDEdit \cite{meng2022sdedit} follows an inversion-based approach for performing guided image synthesis. In particular, the generative prior is introduced by first passing the painting $y$ through the forward diffusion pass $y \rightarrow y_{t_0}$ \cite{kim2022diffusionclip,song2020denoising}, and then performing reverse diffusion $y_{t_0} \rightarrow y_0$ to get the output image $x = y_0$. Due to space constraints, we primarily use the standard hyperparameter value of $t_0=0.8$ in the main paper. In this section, we provide additional results which comprehensively compare our approach with SDEdit \cite{meng2022sdedit} under changing values of $t_0$. 

\begin{figure*}[htbp]
\begin{center}
\centering
     \begin{subfigure}[b]{0.98\textwidth}
         \centering
         \includegraphics[width=\textwidth]{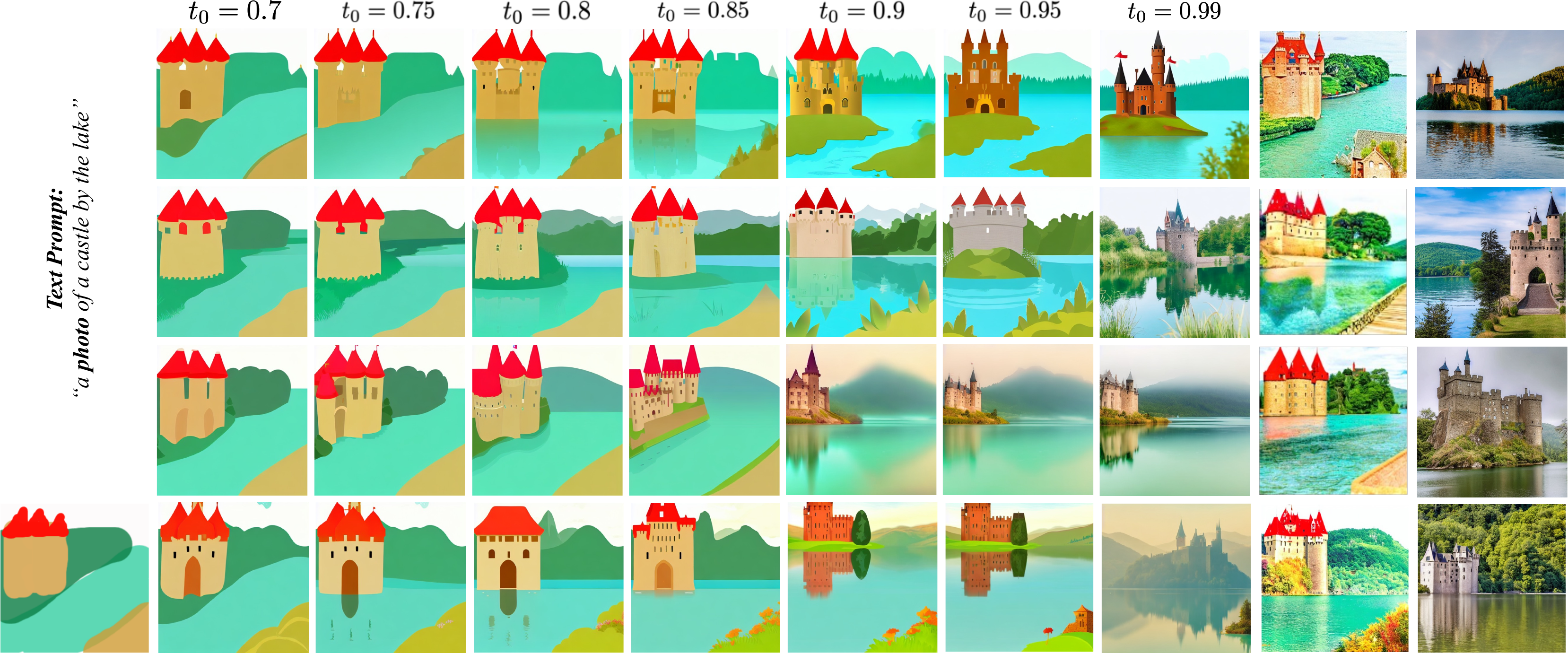}
     \end{subfigure} 
     \vskip 0.02in
      \begin{subfigure}[b]{0.98\textwidth}
         \centering
         \includegraphics[width=\textwidth]{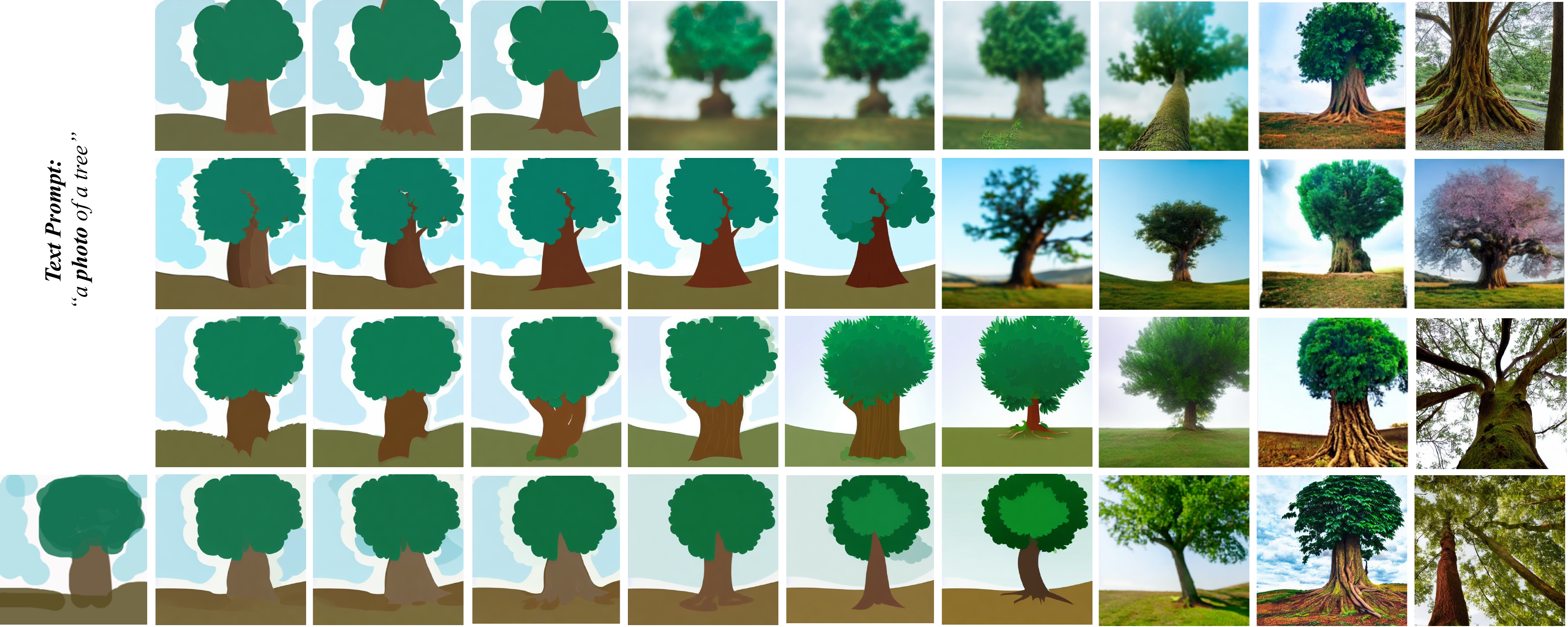}
     \end{subfigure}
     \vskip 0.02in
      \begin{subfigure}[b]{0.98\textwidth}
         \centering
         \includegraphics[width=\textwidth]{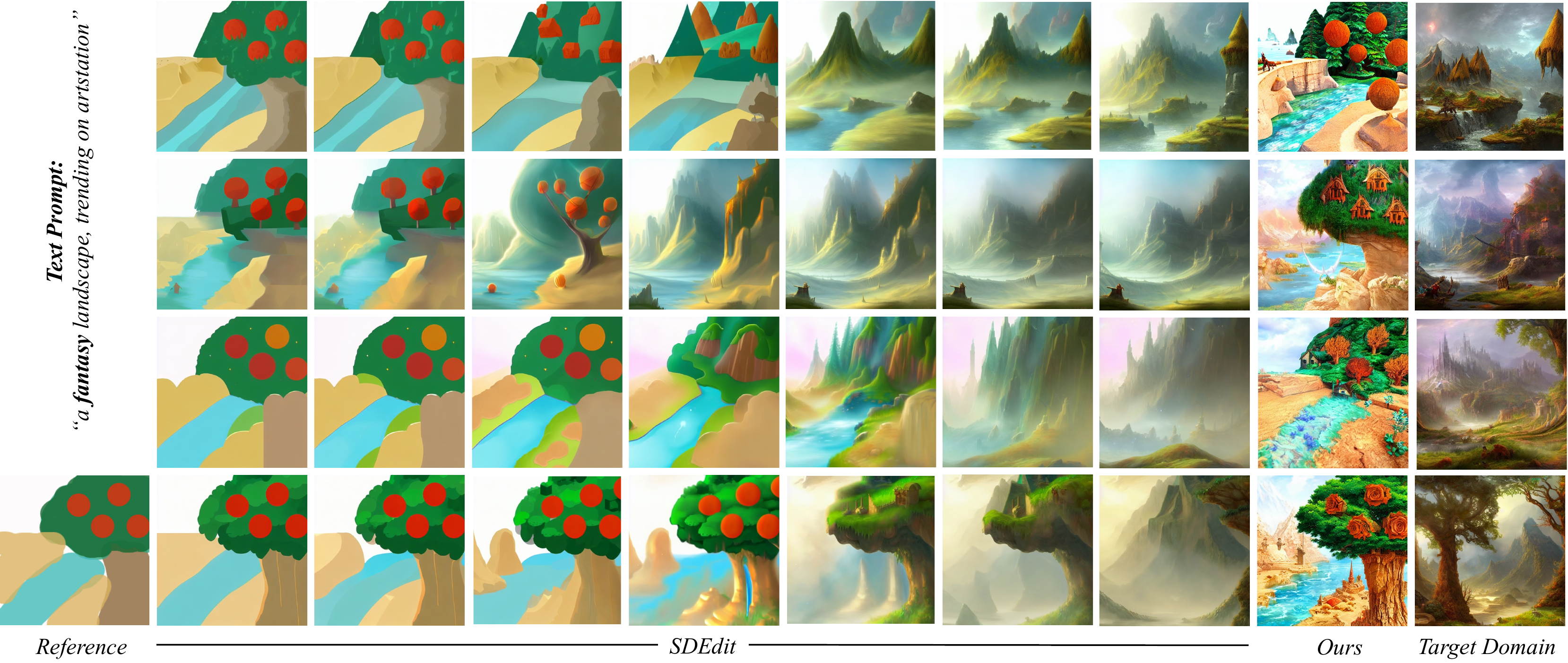}
     \end{subfigure}
\vskip -0.05in
\caption{\emph{\textbf{Additional comparisons.}} We provide comprehensive comparisons with SDEdit \cite{meng2022sdedit} under changing value of hyperparameter $t_0$. We find that SDEdit \cite{meng2022sdedit} either generates faithful but cartoon-like outputs for low $t_0$, or, generates realistic but unfaithful outputs at high $t_0$. In contrast, our approach leads to outputs which are both realistic (\emph{w.r.t} the target domain) as well as faithful (to the provided reference).
}
\label{fig:sdedit-var-p2}
\end{center}
\vskip -0.3in
\end{figure*}

\begin{figure*}[htbp]
\begin{center}
\centering
     \begin{subfigure}[b]{0.98\textwidth}
         \centering
         \includegraphics[width=\textwidth]{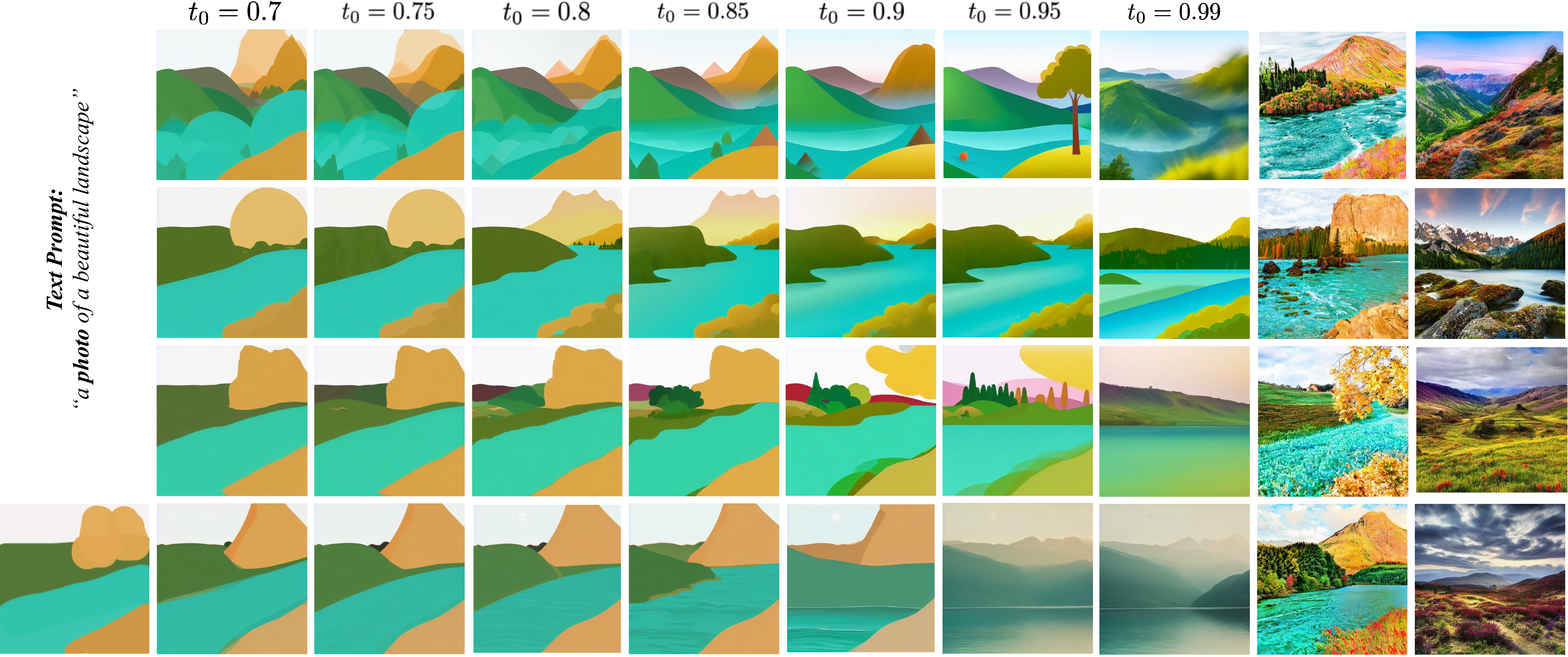}
     \end{subfigure} 
     \vskip 0.02in
      \begin{subfigure}[b]{0.98\textwidth}
         \centering
         \includegraphics[width=\textwidth]{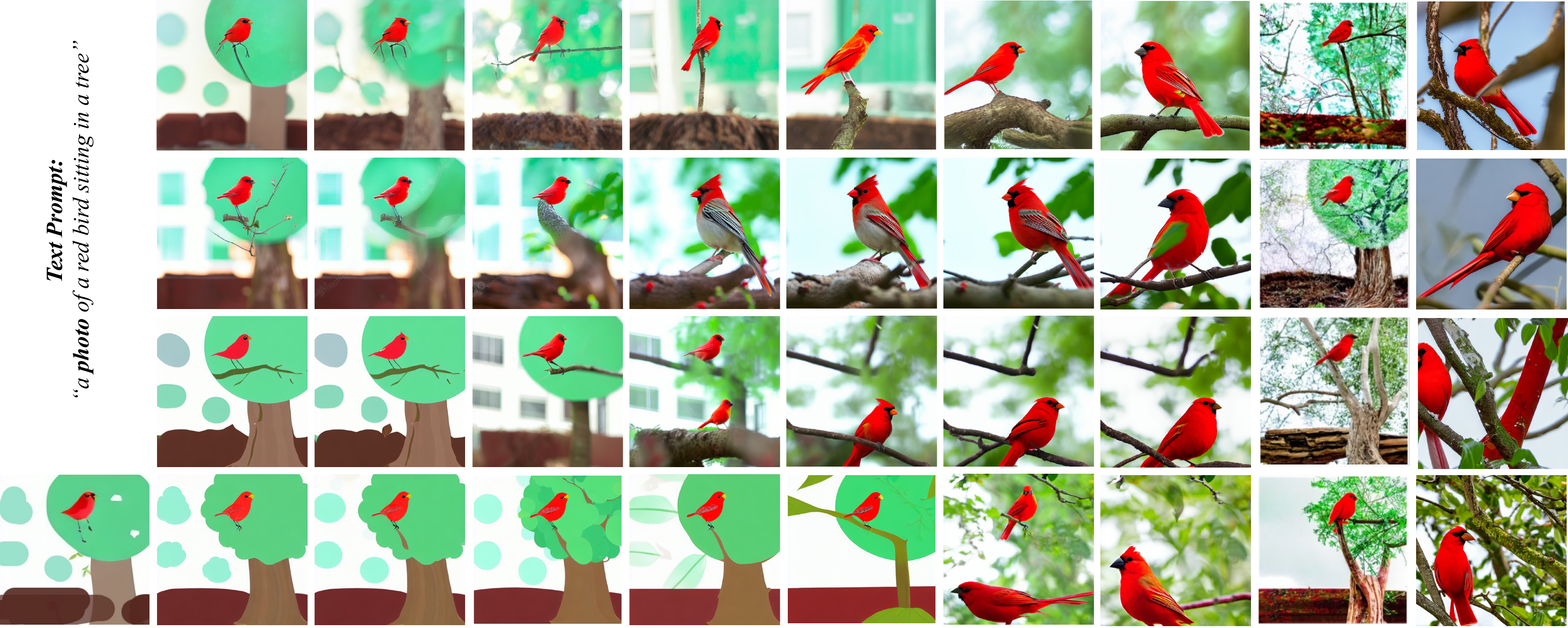}
     \end{subfigure}
     \vskip 0.02in
      \begin{subfigure}[b]{0.98\textwidth}
         \centering
         \includegraphics[width=\textwidth]{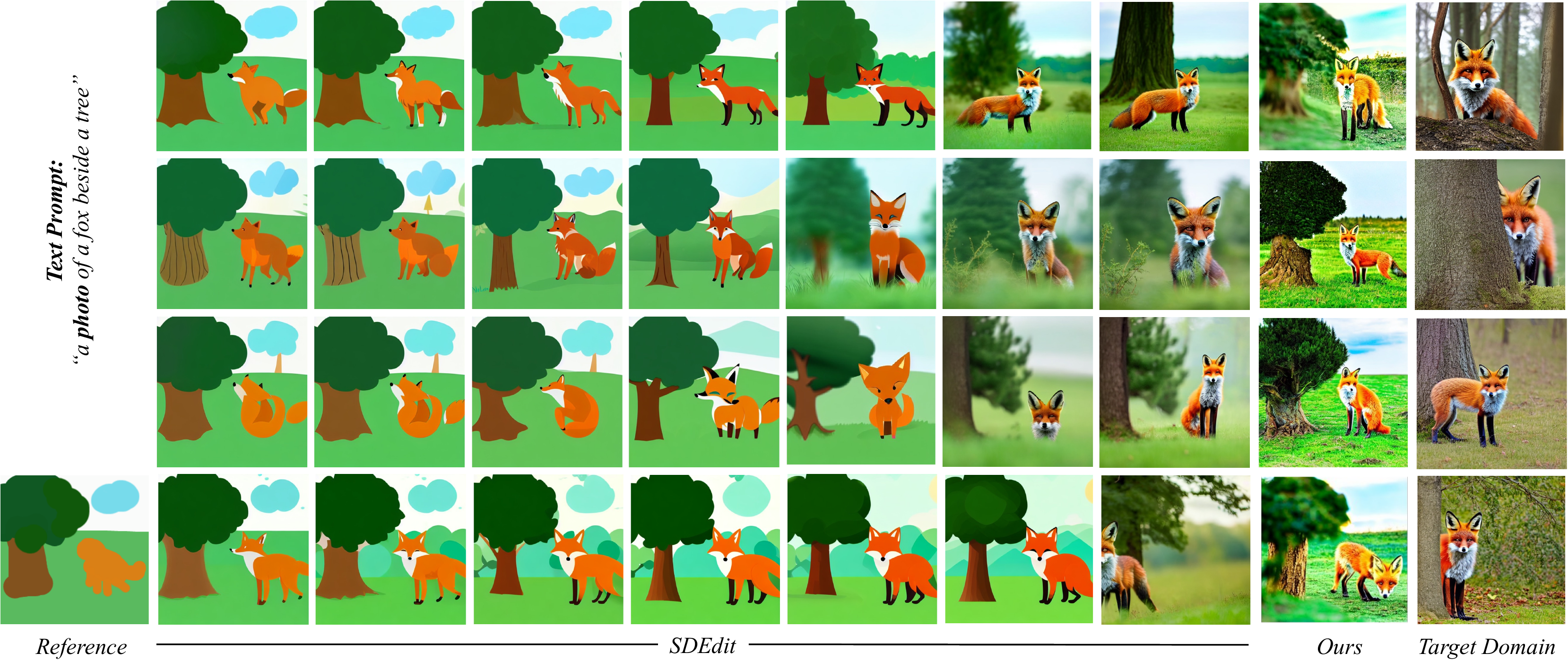}
     \end{subfigure}
\vskip -0.05in
\caption{\emph{\textbf{Additional comparisons.}} We provide comprehensive comparisons with SDEdit \cite{meng2022sdedit} under changing value of hyperparameter $t_0$. We find that SDEdit \cite{meng2022sdedit} either generates faithful but cartoon-like outputs for low $t_0$, or, generates realistic but unfaithful outputs at high $t_0$. In contrast, our approach leads to outputs which are both realistic (\emph{w.r.t} the target domain) as well as faithful (to provided reference).
}
\label{fig:sdedit-var-p1}
\end{center}
\vskip -0.3in
\end{figure*}

\emph{\textbf{Qualitative Comparisons.}} Results are shown in Fig.~\ref{fig:sdedit-var-p2}, \ref{fig:sdedit-var-p1}. We observe that for lower values of $t_0$, SDEdit generates outputs which though highly faithful to the reference painting, lack details and represent simplistic representations of the target domain. Increasing the value of hyperparameter $t_0$ helps improve realism but the outputs become less and less faithful with the reference painting. In contrast, we find that the proposed approach leads to outputs which are both \emph{faithful} to the reference painting as well as exhibit high \emph{realism} \emph{w.r.t} the target domain (which is generated using only text prompt conditioning).

\begin{figure}[t]
\begin{center}
\centerline{\includegraphics[width=0.98\linewidth]{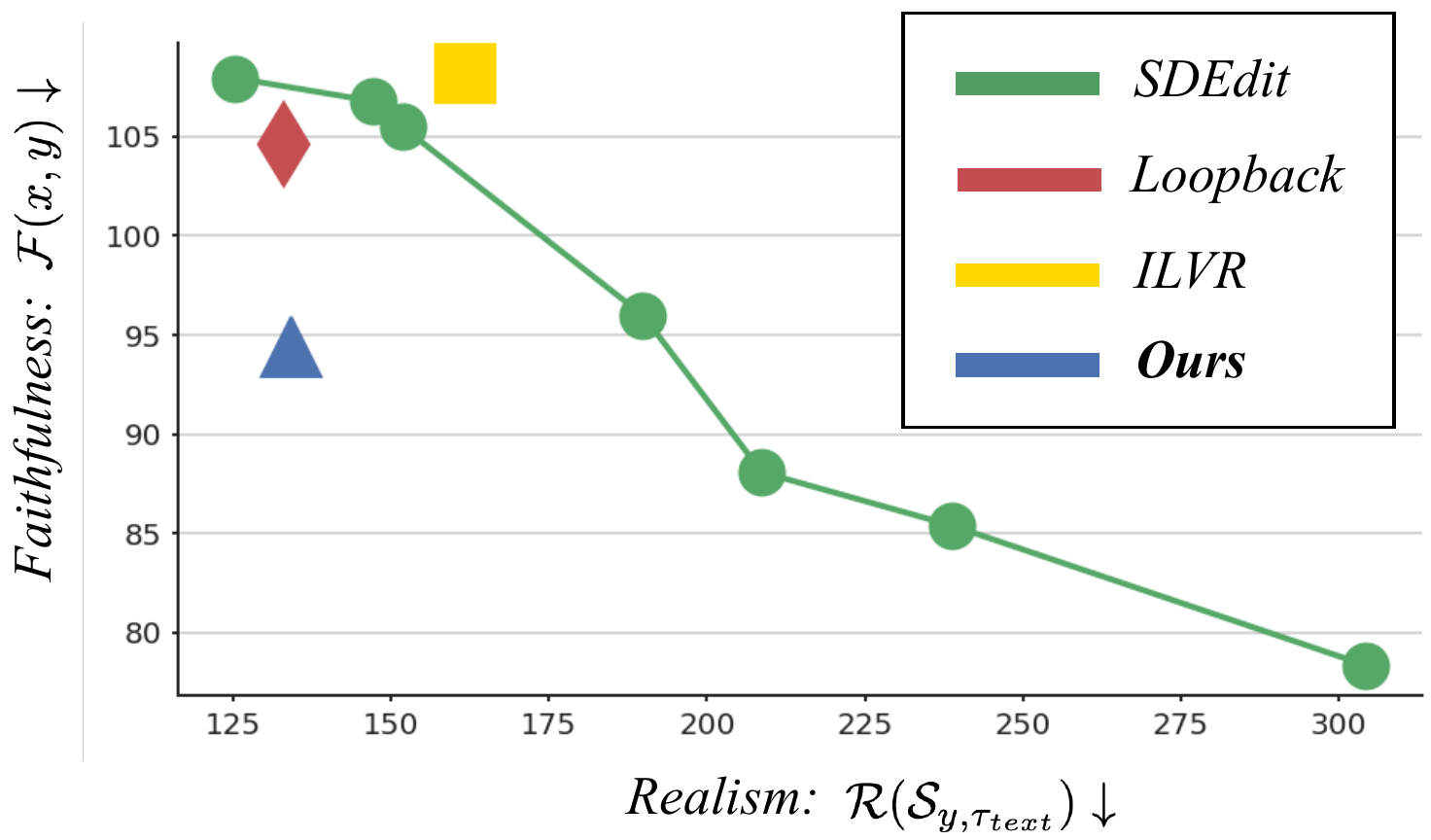}}
\vskip -0.05in
\caption{\emph{\textbf{Visualizing faithfulness-realism tradeoff}.} We analyse the tradeoff between faithfulness-realism distances for different methods (note that lower is better for both metrics). We observe that as compared to prior works, our method provides the best tradeoff between generating realistic outputs and maintaining faithfulness with the provided reference painting.}
\label{fig:quant-results}
\end{center}
\vskip -0.3in
\end{figure}

\emph{\textbf{Quantitative Comparisons.}} In addition to qualitative results, we also report quantitative results by analysing the relationship between the \emph{faithfulness} $\mathcal{F}$ and \emph{realism} $\mathcal{R}$ metrics (refer Sec.~\textcolor{red}{4.1} of main paper), under changing values hyperparameter $t_0$. 
Results are shown in Fig.~\ref{fig:quant-results}. We observe that as compared to prior works, our method provides the best tradeoff generating \emph{realistic} outputs and maintaining \emph{faithfulness} with the provided reference painting.

\subsection{Comparison with Custom Baselines}
\label{sec:custom-baselines}

\begin{figure*}[h!]
\begin{center}
\centering
     \begin{subfigure}[b]{0.98\textwidth}
         \centering
         \includegraphics[width=\textwidth]{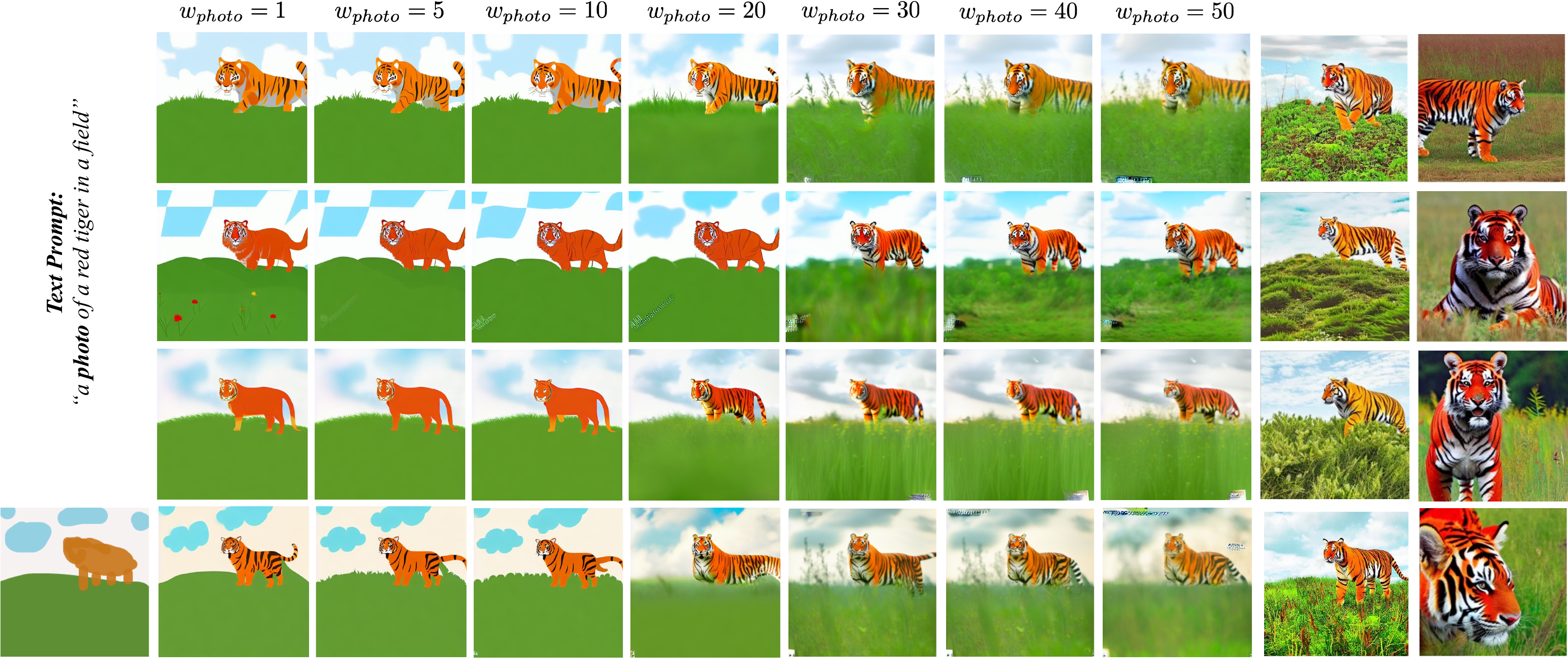}
     \end{subfigure} 
     \vskip 0.02in
      \begin{subfigure}[b]{0.98\textwidth}
         \centering
         \includegraphics[width=\textwidth]{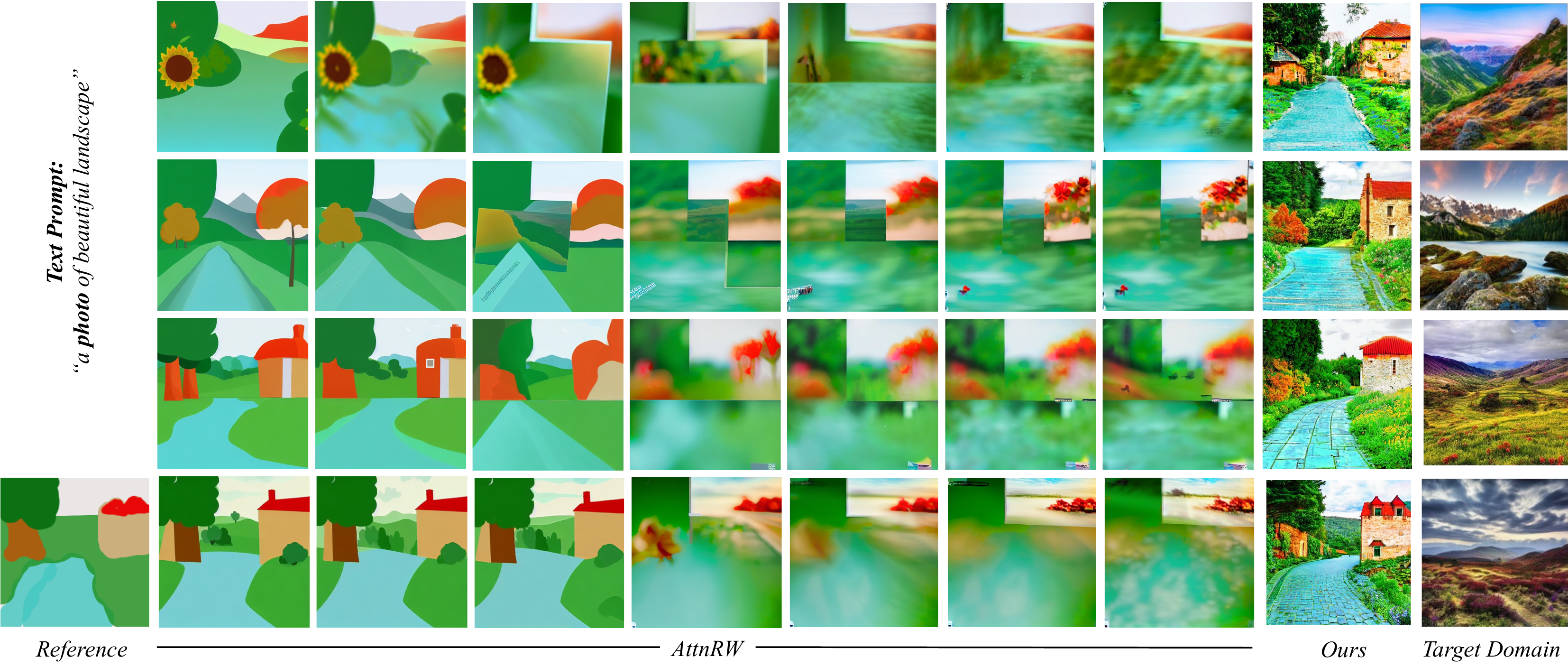}
     \end{subfigure}
\vskip -0.05in
\caption{\emph{\textbf{Comparison with Custom Baselines - AttnRW} \cite{hertz2022prompt}.} We compare the performance of our method with the Attention Reweighting (AttnRW) approach for increasing realism \emph{w.r.t} the target domain. We find that increasing the weight of cross attention maps corresponding to the domain-specific text tokens (\eg photo in above), leads to improved realism of the generated outputs. 
However, we note that certain blurry details persist \eg grass in row 1-4. Also, the increase in realism is accompanied by some image artifacts \eg blotched regions in row 1-4,  image in image artifacts in row 4-8 \etc. In contrast, our approach improves output realism in a more coherent manner.
}
\label{fig:attnrw-var}
\end{center}
\vskip -0.3in
\end{figure*}

\begin{figure*}[h!]
\begin{center}
\centerline{\includegraphics[width=1.\linewidth]{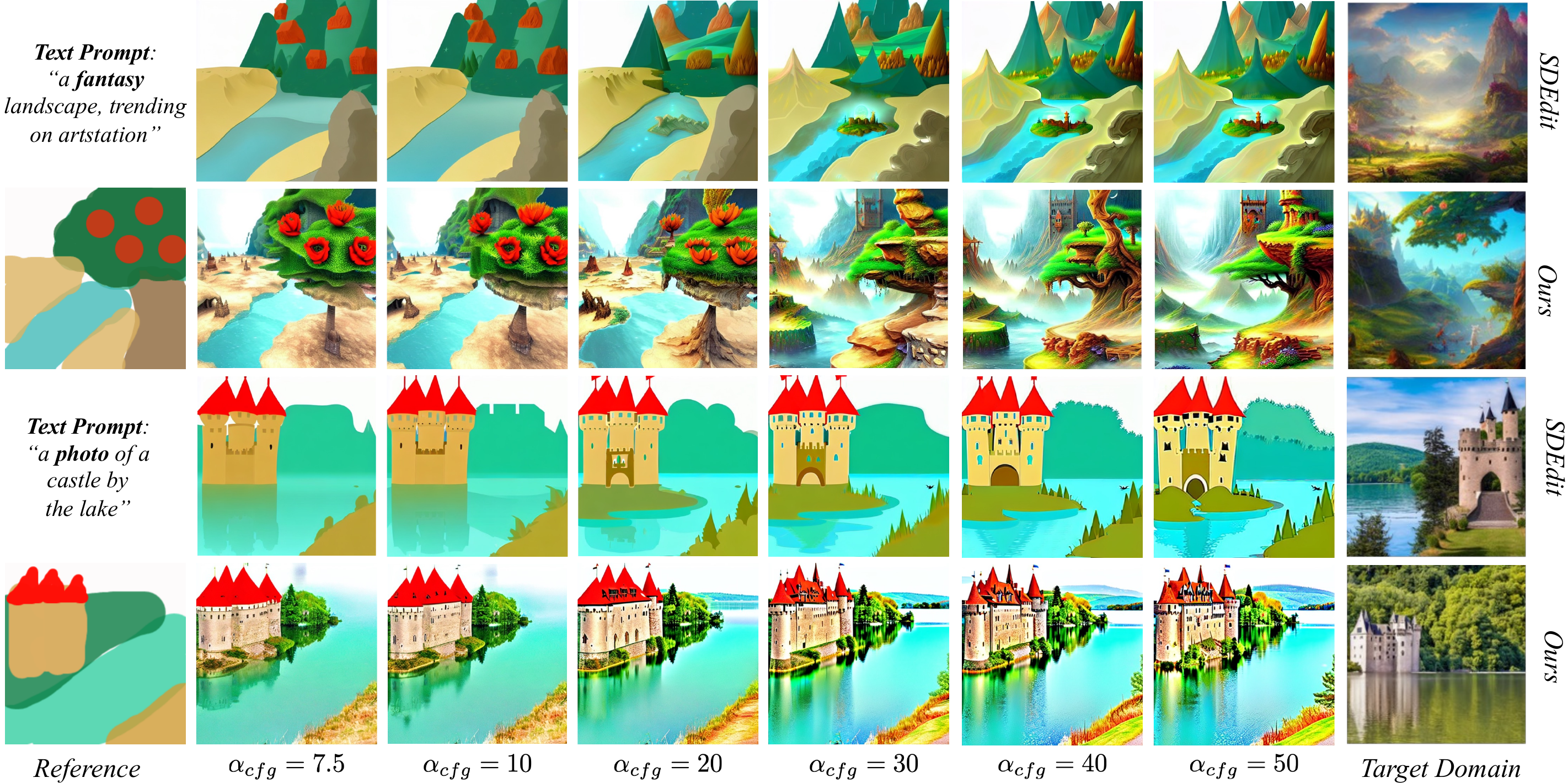}}
\vskip -0.05in
\caption{\emph{\textbf{Comparison with Custom Baselines - CFG}} \cite{ho2022classifier}. We analyse the impact of increasing the classifier-free guidance scale $\alpha_{cfg}$ on outputs generated using SDEdit \cite{meng2022sdedit} and our method. We find that while increasing the value of $\alpha_{cfg}$ leads to increase in level of details, the final outputs still represent simplistic representations of the target domain (row-3). Furthermore, as the value of $\alpha_{cfg}$ is increased, the faithfulness with respect to the reference painting is compromised (\eg red regions in row-1).
}
\label{fig:cfg-var}
\end{center}
\vskip -0.3in
\end{figure*}

In this section, we introduce some custom methods (as baselines) for increasing the realism of generated outputs with SDEdit \cite{meng2022sdedit}, and then compare the output performance for the same with our approach. In particular, we show additional comparisons with the following custom baselines,

\begin{itemize}[topsep=1pt,itemsep=0ex,leftmargin=*]
    \item \emph{\textbf{Attention Re-weighting (AttnRW)}} \cite{hertz2022prompt} wherein the realism \emph{w.r.t} the target domain is enhanced by increasing the attention weighting for the corresponding domain specific text tokens (\eg  photo, painting \etc).
For instance, if the text prompt says \emph{``a photo of a tree''}, then we aim to increase the realism of the generated outputs by increasing the weightage of the cross-attention maps corresponding to the the word \emph{``photo''} \cite{hertz2022prompt}. Results are shown in Fig.~\ref{fig:attnrw-var}. We observe that while increasing the weightage of domain specific text tokens (\eg  photo, painting \etc) helps improve the realism of the output images to some extent, the final images still lack details and certain blurry regions still persist (\eg grass in row-1). Furthermore, the increase in realism is accompanied by some image artifacts \eg blotched image regions in row 1-4,  image-in-image artifacts in row 4-8 \etc. 
In contrast, we find that our method provides a more practical approach for increasing the output realism in a semantically coherent manner.
    \item \emph{\textbf{Increasing Classifier Guidance Scale}} \cite{ho2022classifier}, wherein we attempt to increase the realism of the SDEdit \cite{meng2022sdedit} outputs by increasing the scale of classifier free guidance used during the reverse diffusion process. Results are shown in Fig.~\ref{fig:cfg-var}. We observe that while increasing the scale of classifier free guidance improves the level of detail in the generated images, the final outputs still resemble cartoon-like or simplistic representations of the target domain. Furthermore, we also note that our approach can also benefit from the increase in guidance scale in order to increase the level of fine-grain detail in the output images (\eg details of castle, water reflections in Fig.~\ref{fig:cfg-var}).
\end{itemize}

\section{Experiment Details}
\subsection{Implementation Details}

In this section, we provide further details for the implementation of our approach as well as other baselines used while reporting results in the main paper.

\textbf{Ours.} We use publicly available text-conditioned latent diffusion models \cite{diffusers, rombach2021highresolution} for implementing the purposed approach in the main paper.
The constrained optimization is performed using gradient descent with the Adam \cite{kingma2014adam} optimizer and number of gradient steps $N_{grad}\in[20,60]$.
While several formulations of the distance measure $\mathcal{L}$ and painting function $f$ are possible (refer Sec.~\ref{sec:method-analysis}), we find that simply approximating the function $\mathcal{L}$ using mean squared distance and $f$ as a convolution operation with a gaussian kernel seems to give the fastest inference time performance with our method. For consistency with prior works, we use the non-differentiable painting function from SDEdit \cite{meng2022sdedit} while reporting quantitative results. All results are reported using the DDIM sampling \cite{song2020denoising} with 50 inference steps for performing the reverse diffusion process.

\textbf{SDEdit}\cite{meng2022sdedit}. We use the standard image-to-image pipeline from the open-source \emph{diffusers} library \cite{diffusers} for reporting results for \emph{SDEdit} \cite{meng2022sdedit} with different values of hyperparameter $t_0 \in [0,1]$. Similar to our method, all results are reported at $512 \times 512$ resolution using DDIM sampling \cite{song2020denoising} with 50 inference steps for performing the reverse diffusion process. Unless otherwise specified, a classifier-free guidance scale \cite{ho2022classifier} of $7.5$ is used for all experiments.

\textbf{SDEdit + Loopback}\cite{loopback}. We use the previously described SDEdit implementation and iteratively reperform guided synthesis with the previous diffusion outputs to improve realism of the generated outputs. In particular, we use $N_{iter}=4$ iterations for the iterative process. Also, similar to \cite{loopback}, in order to increase the realism of generated outputs with each iteration, the hyperparameter $t_0$ is updated as,
\begin{align}
    t_0^{n+1} \leftarrow min (t_0^n \ \cdot \  k, 1.0),  \quad k \in [1.0, 1.1]
\end{align}
where $n \in [1,N_{iter}]$ is the iteration number. Unless otherwise specified, we use the standard hyperparameter selection of $k=1.05$ and $t^{n=1}_0=0.8$ for our experiments.

\textbf{ILVR} \cite{choi2021ilvr}. The original ILVR  \cite{choi2021ilvr} algorithm was proposed for iterative refinement with diffusion models in pixel space. We adapt the ILVR implementation for inference with latent diffusion models \cite{rombach2021highresolution} for the purposes of this paper. In particular given a reference painting $y$, the original ILVR algorithm modifies the diffusion output $x_t$ (in pixel space) at any timestep $t$ during reverse diffusion process as, 
\begin{align}
    \Tilde{x}_t = \phi_N(y_t) + x_t - \phi_N(x_t), \quad y_t \sim q(y_t\mid y)
\end{align}
where $q(y_t\mid y)$ represents the forward diffusion process from $y \rightarrow y_t$, $\phi_N(.)$ is a low pass filter achieved by scaling down the image by a factor of $N$ and then upsampling it back to the original dimensions. Assuming a latent diffusion model with encoder $\mathcal{E}$ and decoder $\mathcal{D}$, we simply adapt the above update in latent space as follows,
\begin{align}
   x_t = \mathcal{D}(z_t) \label{eq:ilvr-z-1}\\
   z_y =  \mathcal{E}(y), \ z_{y_t} \sim q(z_{y_t} \mid z_{y})\label{eq:ilvr-z-4}\\
   \Tilde{x}_t = \phi_N(y_t) + x_t - \phi_N(x_t), \quad y_t = \mathcal{D}(z_{y_t}) \label{eq:ilvr-z-2}\\
   \Tilde{z}_t = \mathcal{E}(\Tilde{x}_t) \label{eq:ilvr-z-3}
\end{align}
where Eq.~\ref{eq:ilvr-z-1}, \ref{eq:ilvr-z-3} map the latent features $z_t$ to pixel space $x_t$, and vice-versa. Eq.~\ref{eq:ilvr-z-4} computes $y_t$ from $y$  by first mapping $y$ to $z_y$, computing the forward diffusion $z_{y} \rightarrow z_{y_t}$ and then reverting back $z_{y_t}$ to $y_t$. Finally, Eq.~\ref{eq:ilvr-z-2}  is simply the original update rule from ILVR algorithm \cite{choi2021ilvr}.
A hyperparmeter value of $N=4$ is used while reporting results.


\subsection{Quantitative Experiments}
\label{sec:quant-experiments}

\textbf{Data Collection}. Since there is no predefined dataset for guided image synthesis with user-scribbles and text prompts, we create our own dataset for reporting quantitative results. In particular, we first collect a set of 100 stroke painting and text prompt pairs from diverse data modalities with the help of actual human users. We then augment the collected data using a prompt-engineering approach to increase the diversity of the collected data pairs. In particular, the text prompt for each data-pair is modified in order to replace the
domain specific text words (\eg photo, painting) with pre-designed target domain templates, while keeping the underlying content the same.
During prompt engineering, the target domain template is chosen randomly from \emph{[  `photo',`watercolor painting', `Vincent Van Gogh painting',`children drawing',`high resolution disney scene', `high resolution anime scene', `fantasy scene',`colored pencil sketch']}. 
For each data pair, we then sample four random guided image synthesis outputs for each baseline and our method.
The resulting dataset consists of 800 (painting, text-prompt) pairs and 3200 overall samples from diverse data modalities for final method evaluation.

\begin{figure}[t]
\begin{center}
\centerline{\includegraphics[width=1.\linewidth]{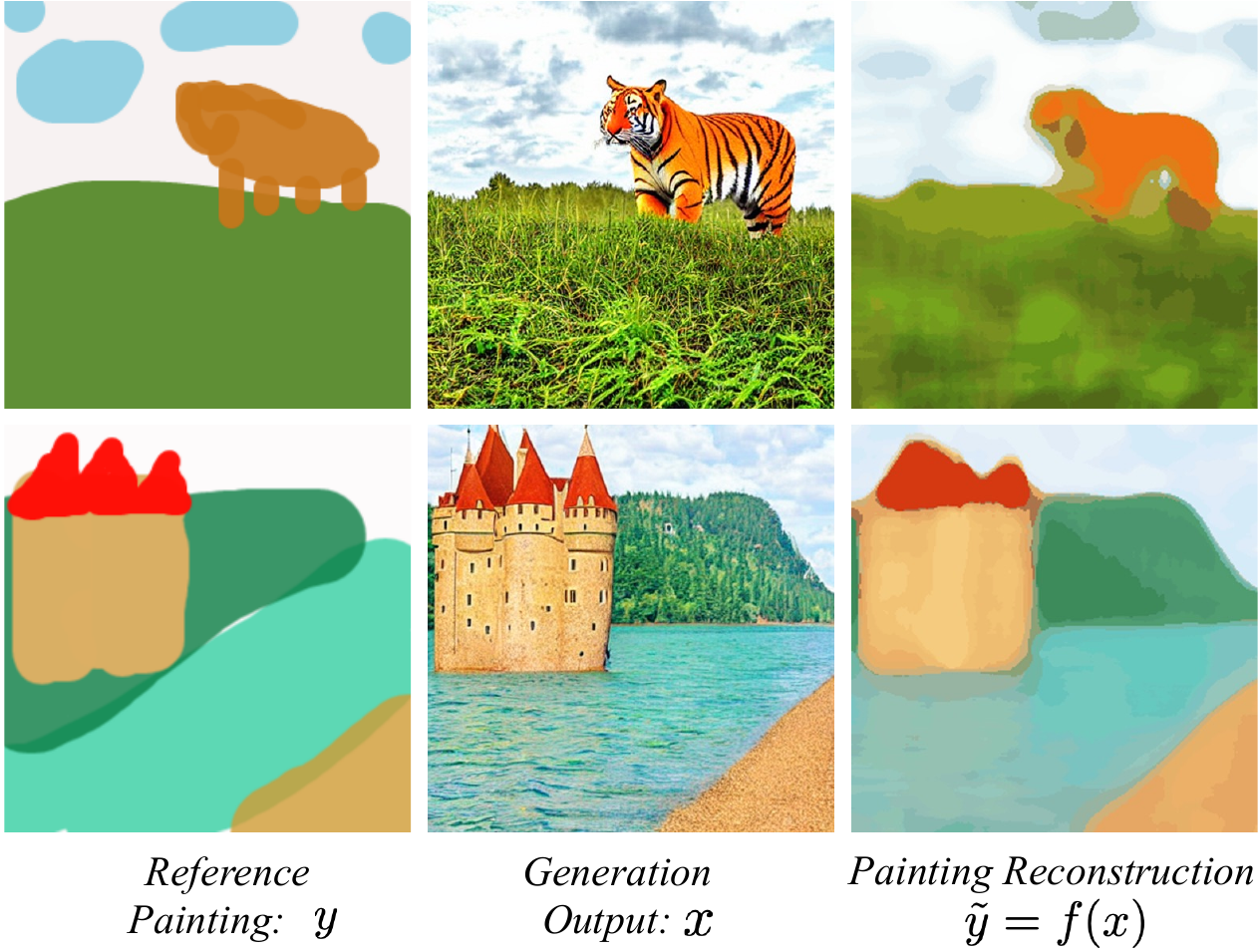}}
\vskip -0.05in
\caption{Visualizing input painting $y$, output $x$ and painted reconstruction $\Tilde{y}=f(x)$. The goal is to generate an output $x$ which is realistic and for which painting loss $\mathcal{L}_2(f(x),y)$ is minimized.}
\label{fig:painting-function-vis}
\end{center}
\vskip -0.3in
\end{figure}

\textbf{Quantitative Metrics.} In order to evaluate the performance of our approach, we introduce two metrics for measuring the \emph{faithfulness} of the output \emph{w.r.t} the reference painting, and the \emph{realism} of the generated samples \emph{w.r.t} the target domain (specified through text-only conditioning). In particular, given an input painting $y$ and output real image prediction $x$, we define faithfulness distance $\mathcal{F}(x,y)$ as,
\begin{align}
    \mathcal{F}(x,y) = \mathcal{L}_2(f(x),y)
\end{align}
where $f(.)$ is the painting function. Thus an output image $x$ is said to have high faithfulness with the given painting $y$ if upon painting the final output $x$ we get a painting $\Tilde{y}=f(x)$ which is similar to the original target painting $y$ (Fig.~\ref{fig:painting-function-vis}).

The painting function $f$ is implemented using the human stroke-simulation algorithm from SDEdit \cite{meng2022sdedit}. In particular, given an $256 \times 256$ input image, the output painting is computed by first passing the image through a median filter with kernel size $23$, and then perform color quantization to reduce the number of colors to $20$ using an adaptive palette.

\begin{figure*}[h!]
\begin{center}
\centerline{\includegraphics[width=1.\linewidth]{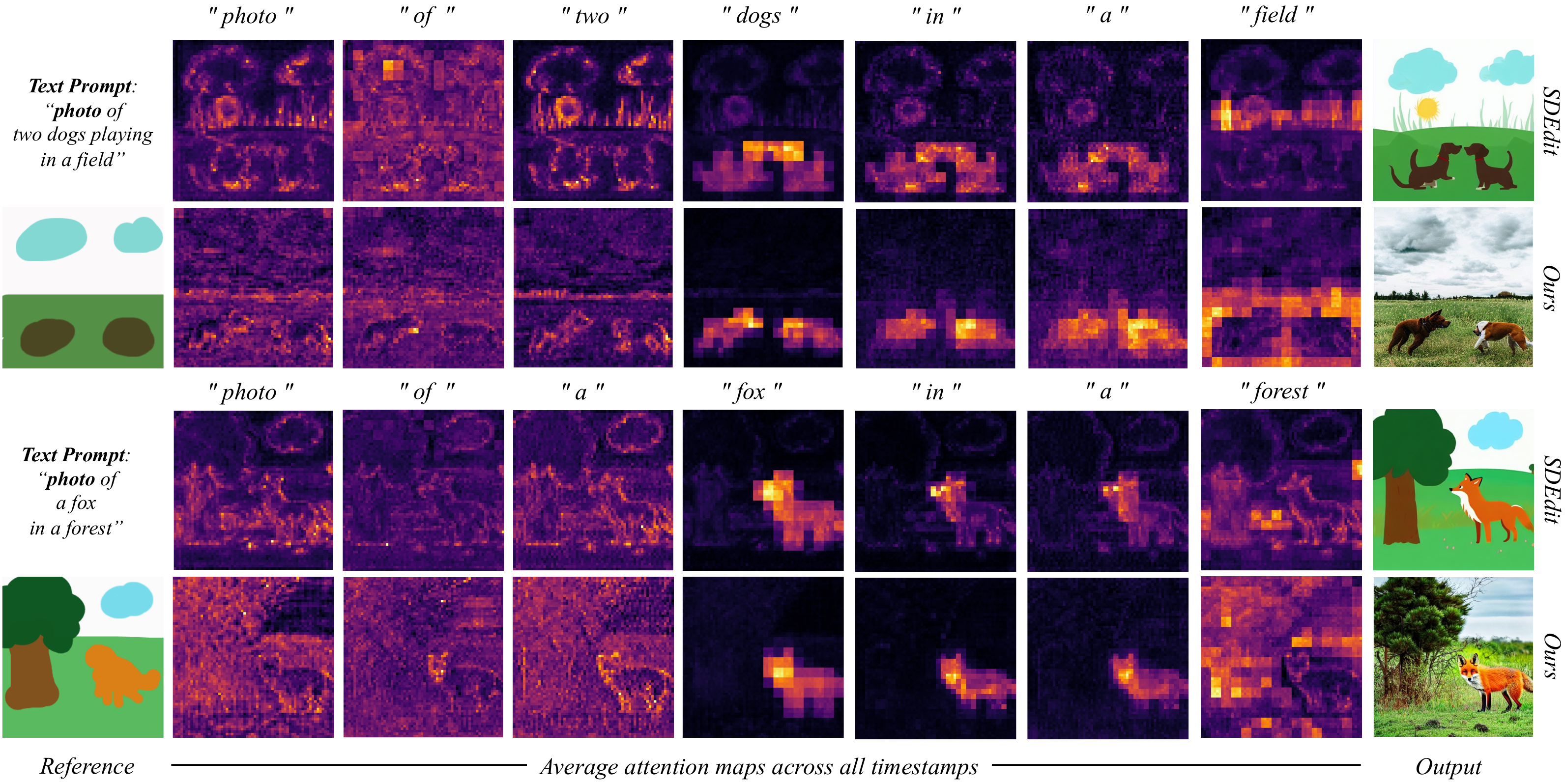}}
\vskip -0.05in
\caption{\emph{\textbf{Visualizing the effect of GradOP on cross attention maps.}}  We analyse the effect of our approach on the cross-attention maps generated during the reverse diffusion process. We find that our method leads to cross-attention outputs which help the model pay better attention to desired image areas in the reference painting. For instance, in the first example, the cross-attention features show high overlap with the desired \emph{dog} and \emph{field} regions. In contrast, the cross attention maps from SDEdit \cite{meng2022sdedit} reveal that the model is not paying adequate attention to the desired image areas (\eg \emph{field} in  row-1, \emph{tree and forest} in row-3) while generating the final output.
}
\label{fig:cross-attn-vis}
\end{center}
\vskip -0.3in
\end{figure*}

Similarly, given a set of output data samples $\mathcal{S}({y,\tau_{text}})$ conditioned on both painting $y$ and text $\tau_{text}$, and, $\mathcal{S}({\tau_{text}})$ conditioned only on the text, the \emph{realism} $\mathcal{R}$ is defined as,
\begin{align}
    \mathcal{R}(\mathcal{S}({y,\tau_{text}})) = FID \left(\mathcal{S}({y,\tau_{text}}),\mathcal{S}({\tau_{text}})\right)
\end{align}
where $FID$ represents the Fisher inception distance \cite{heusel2017gans}. 

Please note that while the above defined \emph{realism} distance measure $\mathcal{R}$ captures the realism with respect to the target domain, we expect the computed $FID$ scores to be higher than those expected of unconditioned image outputs. This is because while the proposed method generates outputs which seem realistic to human eyes, the variance of output distribution is significantly lower than that of real images. The decreased variance in output images occurs simply because the layout and color composition are predominantly fixed as a result of additional conditioning on the stroke painting $y$. In contrast, natural images or images conditioned only on the text prompt have a much higher diversity in terms of scene layout and the overall color composition. We therefore try to overcome of lack of diversity in generated image outputs by performing random data augmentations (random horizontal flip and random resized crop of size $448 \times 448$ on a $512 \times 512$ image) while computing the final realism scores 
across different methods \footnote{Note that while this helps increase the diversity in scene layout the diversity in color composition is still lower than that of real images or image outputs conditioned only on the text prompt.}.

\textbf{Human User Study}. In addition to reporting quantitative results using the above defined measures for \emph{faithfulness} and \emph{realism}, similar to \cite{meng2022sdedit}, we also perform a human user study wherein the \emph{realism} and the overall satisfaction score (\emph{faithfulness} + \emph{realism}) are evaluated by actual human users. 
For the \emph{realism} scores, given an input text prompt (with target domain $\tau_{domain}$ \eg $\tau_{domain} = $ `\emph{photo}') and sample images conditioned only on the text prompt, the participants were shown a pair of image generation outputs comparing our method with prior works. For each pair, the human subject is then asked to select the output image which is more realistic with respect to the target domain ($\tau_{domain}$). 
Similarly, for computing the overall satisfaction scores,  given an input stroke painting, text prompt and sample images conditioned only on the text prompt, the participants were shown a pair of image generation outputs comparing our method with prior works. For each pair, the instruction is: ``Given the input painting and text prompt, how would you imagine this image to look like in reality? 
Your selection should be based on
how realistic and less blurry the image is (please check level of details), consistency with the target domain ($\tau_{domain}$) and whether it is faithful with the reference painting in terms of scene layout, color composition''. For each task (\eg computing overall satisfaction score), the collected data samples (discussed above) were divided among 50 human participants, who were given an unlimited time in order to ensure high quality of the final results. Additionally, in order to remove data noise, we use a repeated comparison (control seed) for each user. Responses of users who answer differently to this repeated seed are discarded while reporting the final results.

\section{Method Analysis: Continued}
\label{sec:method-analysis}

\begin{figure*}[t]
\begin{center}
\centerline{\includegraphics[width=1.\linewidth]{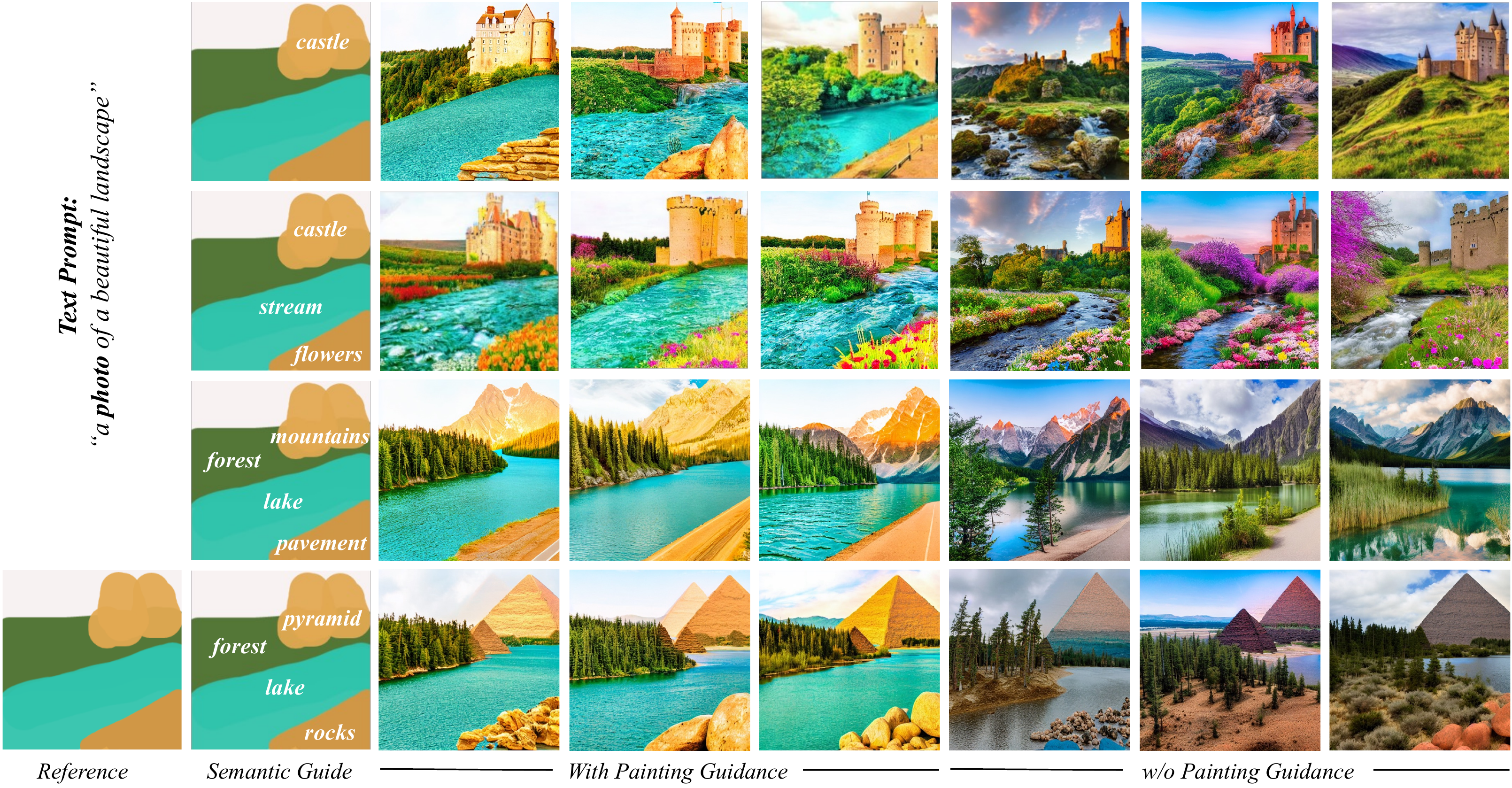}}
\vskip -0.05in
\caption{\emph{\textbf{Analysing role of painting guidance in semantic control.}}  We analyse the effect of using an underlying reference painting as guidance in controlling the semantics of different image areas using cross-attention based correspondence approach presented in the main paper (refer Sec.~\textcolor{red}{3.3} in main paper). We find that additional guidance using reference stroke painting helps the user gain much accurate control over the semantics of different image regions (\eg lake in row-3,4, mountains in row-3, rocks, forest in row-4 \etc).
}
\label{fig:paint-guid-sem-control}
\end{center}
\vskip -0.3in
\end{figure*}

\subsection{Effect of GradOP on Cross Attention Maps}

As shown by Hertz \etal \cite{hertz2022prompt} and our results, the cross-attention maps corresponding to different words in the input text prompt play a key role in deciding the overall semantic contents of the final image output. In this section, we try to analyse how the proposed approach leads to more realistic image content generation by analysing the average cross-attention maps generated while performing the reverse diffusion process with SDEdit \cite{meng2022sdedit} and our method.

Results are shown in Fig.~\ref{fig:cross-attn-vis}. We find that our method leads to cross-attention outputs which help the model pay better attention to desired image areas in the reference painting. For instance, in the first example, the cross-attention features show high overlap with the desired \emph{dog} and \emph{field} regions. In contrast, the cross attention maps from SDEdit \cite{meng2022sdedit} reveal that the model is not paying adequate attention to the some desired image areas (\eg \emph{field} in  row-1, \emph{forest} in row-3) while generating the final output.

\subsection{Semantic Control without Painting Guidance}

Recall that in addition to performing \emph{high-fidelity} guided image synthesis, we also show that by simply defining a cross attention based correspondence between the input text tokens and the user painting, it is possible to control the semantics of different image regions without the need for any semantic segmentation based conditional training. In this section, we analyse whether similar semantic control is possible without having additional guidance through a stroke painting. In particular, we wish to analyse if such fine-grain control is only possible while providing additional guidance through the reference stroke painting?

To answer this question, we compare the outputs generated through semantic control with and without using a reference painting for the guided synthesis process.
Results are shown in Fig.~\ref{fig:paint-guid-sem-control}. We observe that while for it is feasible to define the semantics of one or two parts of the image accurately using \emph{cross-attention} correspondence, the performance decreases as the number of semantic labels increases (\eg lake in row-3,4, mountains in row-3, rocks, forest in row-4 \etc). In contrast, we find that the use of a reference painting results in much better control over the semantics of different image regions. We believe that the same is because the use of a reference painting sets up a generic semantic structure for the output image which can then be easily refined by defining a cross-attention based correspondence. For instance, in row-4 of Fig.~\ref{fig:paint-guid-sem-control}, adding the blue strokes for lake region sets up a semantic prior which constrains the inference of output semantics to semantic categories like river, lake, sea, stream, blue-green grass, blue pavement \etc. The use of semantic correspondence then helps refine these output semantics to what is actually desired by the user. In contrast, without stroke guidance, the initial semantics for lake region could me much more diverse (\eg sand, rocky terrain in row-4), and thereby more challenging to refine through the proposed semantic correspondence strategy.

\subsection{Inference Time Analysis}
\label{sec:inference-time}

We report a comparison of the average inference times required for each output image in Tab.~\ref{tab:inference-time}. All results are reported using the DDIM sampling \cite{song2020denoising} with 50 inference steps, 
on a single Nvidia RTX 3090 GPU.

\begingroup
\setlength{\tabcolsep}{5.0pt}
\small
\begin{table}[h!]
\begin{center}
\small
\begin{tabular}{l|c|c}
\toprule
\multirow{2}{*}{Method} & \multicolumn{2}{c}{Inference Time (s)} \\
\cline{2-3} 
& \emph{w/o mixed precision} & \emph{with mixed precision} \cite{micikevicius2017mixed}\\
\hline
SDEdit \cite{meng2022sdedit} & 6.32 s & 4.45 s\\
Loopback \cite{loopback} & 27.2 s & 20.46 s\\
ILVR \cite{choi2021ilvr} & 8.24 s & 6.17 s\\
GradOP \textbf{(Ours)} & 20.1 s & 15.8 s\\ 
GradOP+ \textbf{(Ours)} & 12.3 s & 8.86 s\\
\bottomrule
\end{tabular}
\end{center}
\vskip -0.2in
\caption{\emph{\textbf{Inference time analysis}}. Comparing inference time required for generating each output image for different methods. All results are reported with DDIM sampling and 50 inference steps.}
\label{tab:inference-time}
\vskip -0.15in
\end{table}
\endgroup

\subsection{Variation in Painting Function}



Please recall that a key requirement for solving the proposed constrained optimization in Sec.~\textcolor{red}{3} is to define a differentiable painting function $f$, which provides a good approximation for \emph{``how a human would paint a given image with coarse user-scribbles''}.  In this section, we therefore look at some possible formulations for obtaining an approximation of the painting function in a differentiable manner, and compare the corresponding output results.

\begin{figure}[t]
\begin{center}
\centerline{\includegraphics[width=1.\linewidth]{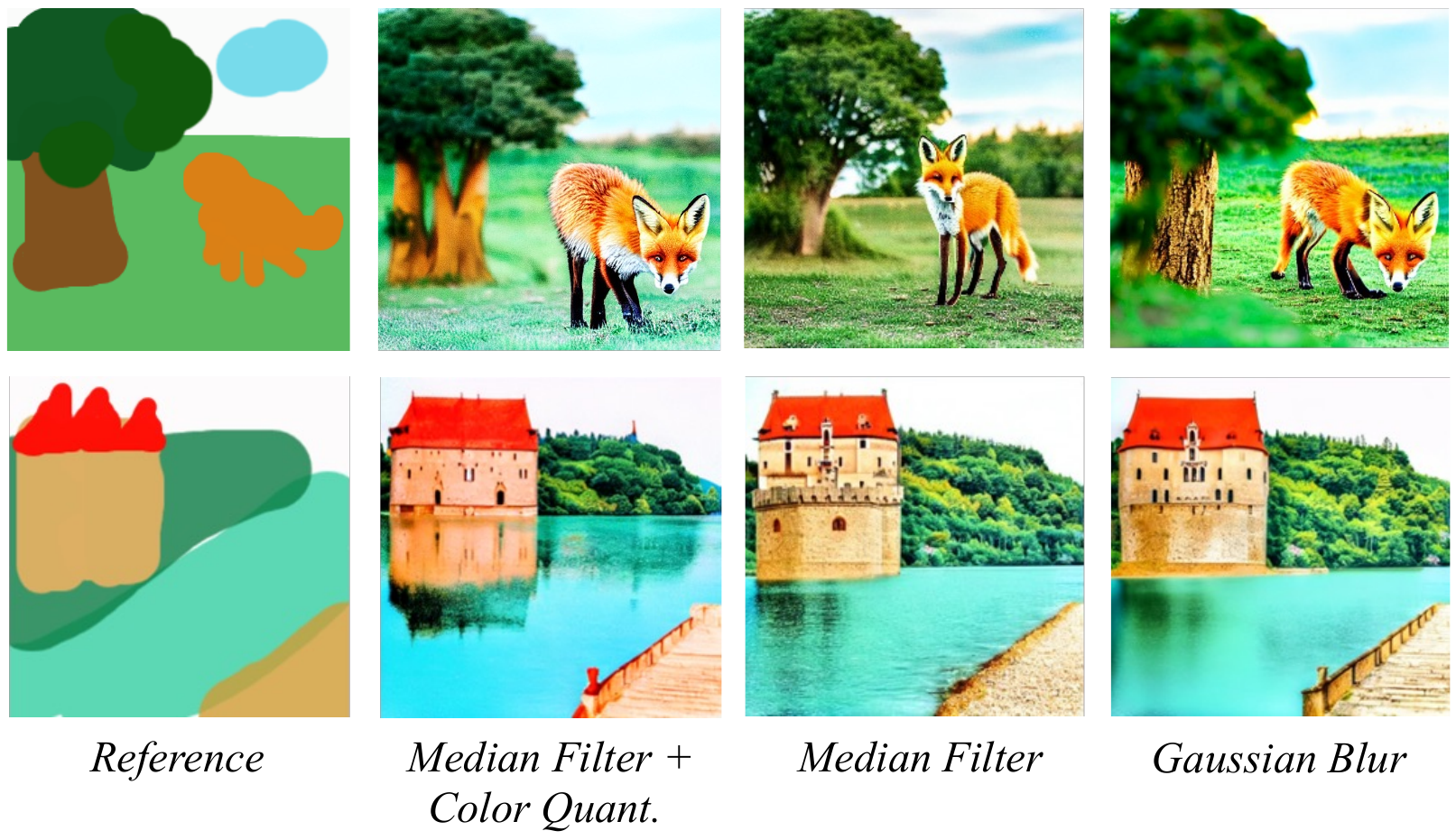}}
\vskip -0.05in
\caption{Analysing performance for different differentiable approximations of the painting function $f$. We find that while using a more accurate painting function \cite{meng2022sdedit} (Col-2) leads to slightly more details (\eg notice the gradient of the grass regions in row-1, detailed shadows of the castle and island in row-2), in practice more simpler approximations (\eg Gaussian Blur) also produces highly realistic outputs while allowing for much faster inference times.
}
\label{fig:paint-func-var}
\end{center}
\vskip -0.3in
\end{figure}

\emph{\textbf{Painting Function Formulation}.} In particular, we consider three main formulations for constructing a differentiable painting function $f$, 1) \emph{Median Filter + Color Quantization}, wherein we implement a differentiable approximation of the human-stroke simulation algorithm in \cite{meng2022sdedit}. In particular, given a reference painting $y$ and output $x$, we first pass $x$ through a median filter of size 23. We then pass the output of the last step through a differentiable color quantization function which maps the image pixels to their nearest $rgb$ value in the painting $y$ (that is, we are performing color quantization \emph{w.r.t} the palette of the reference painting.)
2) \emph{Median Filter} wherein we use the median filter alone for approximating the painting function, and 3) \emph{Gaussian Blur} wherein approximate the painting function through a convolution operation with a Gaussian kernel (size 31 and $\sigma$=7).

Results are shown in Fig.~\ref{fig:paint-func-var}. We observe that while the use of a more accurate human-stroke simulation function from \cite{meng2022sdedit} allows for the generation of slightly more detailed outputs (\eg notice the gradient of the grass regions in row-1, detailed shadows of the castle and island in row-2), it increases the overall inference time required for the proposed gradient descent optimization (40.7s on \emph{GradOP+}). In contrast, we find that using much more simpler approximations (\eg Median Filter, Gaussian Blur) for the painting function also produces highly realistic outputs while allowing for much faster inference times (8.86s, 14.1s on \emph{GradOP+} for Gaussian Blur and Median Filter respectively).

\end{document}